\documentclass[authoryear,a4paper,11pt]{elsarticle_WP}

\usepackage{hyperref}
\usepackage{graphicx}
\usepackage{amssymb,amsmath}
\usepackage{bm}
\usepackage{color}
\usepackage[margin=1in]{geometry}
\usepackage{natbib}
\usepackage{multirow}
\usepackage{booktabs}
\usepackage{setspace}
\usepackage{array}
\usepackage[normalem]{ulem}
\usepackage{tikz}
\usepackage{url}
\usepackage{standalone}
\usepackage{mathtools}
\usepackage{framed}
\newcounter{algorithm}

\journal{Spatial Statistics}

\newcolumntype{x}[1]{>{\centering\arraybackslash\hspace{0pt}}p{#1}}

\newcommand{\zerob} {{\bf 0}}

\newcommand{\thetab} {{\boldsymbol{\theta}}}
\newcommand{\alphab} {{\boldsymbol{\alpha}}}

\newcommand{\nub} {{\boldsymbol{\nub}}}

\newcommand{\etab} {{\boldsymbol{\eta}}}

\newcommand{\varthetab} {{\boldsymbol{\vartheta}}}
\newcommand{\Tset} {\mathcal{T}}
\newcommand{\Yset} {\mathcal{Y}}

\newcommand{\intd} {\textrm{d}}
\newcommand{\phib} {\boldsymbol{\phi}}

\newcommand{\psib} {\boldsymbol{\psi}}

\newcommand{\Sigmamat} {{\bm \Sigma}}

\newcommand{\Amat} {\textbf{A}}

\newcommand{\Kmat} {\textbf{K}}

\newcommand{\Imat} {\textbf{I}}

\newcommand{\Hmat} {\textbf{H}}
\newcommand{\Fmat} {\textbf{F}}

\newcommand{\wvec} {\textbf{w}}
\newcommand{\vvec} {\textbf{v}}
\newcommand{\svec} {\textbf{s}}
\newcommand{\uvec} {\textbf{u}}
\newcommand{\gvec} {\textbf{g}}

\newcommand{\Phimat} {\mathbf{\Phi}}

\newcommand{\nablab} {\boldsymbol{\nabla}}

\renewcommand{\zerob}{\mathbf{0}}

\newcommand{\x}{\mathbf{x}}

\newcommand{\Yvec}{\mathbf{Y}} 

\newcommand{\Zvec}{\mathbf{Z}}
\newcommand{\epsilonb}{\boldsymbol{\varepsilon}}

\newcommand{\var}{\mathrm{var}}

\newcommand{\Gau}{\mathrm{Gau}}

\DeclareMathOperator*{\ext}{ext}
\DeclareMathOperator*{\intr}{int}

\let\originalleft\left
\let\originalright\right
\renewcommand{\left}{\mathopen{}\mathclose\bgroup\originalleft}
\renewcommand{\right}{\aftergroup\egroup\originalright}

\begin{document}

\begin{frontmatter}

\title{\bf Deep Integro-Difference Equation Models for Spatio-Temporal Forecasting}

\author[Wollongong]{Andrew Zammit-Mangion\corref{correspondingauthor}}
\cortext[correspondingauthor]{Corresponding author}
\ead{azm@uow.edu.au}

\author[Missouri]{Christopher K.~Wikle}
\ead{wiklec@missouri.edu}
\address[Wollongong]{School of Mathematics and Applied Statistics, University of Wollongong, Australia}
\address[Missouri]{Department of Statistics, University of Missouri, USA}

\begin{abstract}
Integro-difference equation (IDE) models describe the conditional dependence between the spatial process at a future time point and the process at the present time point through an integral operator. Nonlinearity or temporal dependence in the dynamics is often captured by allowing the operator parameters to vary temporally, or by re-fitting a model with a temporally-invariant linear operator in a sliding window. Both procedures tend to be excellent for prediction purposes over small time horizons, but are generally time-consuming and, crucially, do not provide a global prior model for the temporally-varying dynamics that is realistic.  Here, we tackle these two issues by using a deep convolution neural network (CNN) in a hierarchical statistical IDE framework, where the CNN is designed to extract process dynamics from the process' most recent behaviour. Once the CNN is fitted, probabilistic forecasting can be done extremely quickly online using an ensemble Kalman filter with no requirement for repeated parameter estimation. We conduct an experiment where we train the model using 13 years of daily sea-surface temperature data in the North Atlantic Ocean. Forecasts are seen to be accurate and calibrated. A key advantage of our approach is that the CNN provides a global prior model for the dynamics that is realistic, interpretable, and computationally efficient. We show the versatility of the approach by successfully producing 10-minute nowcasts of weather radar reflectivities in Sydney using the same model that was trained on daily sea-surface temperature data in the North Atlantic Ocean.
\end{abstract}


\end{frontmatter}


\section{Introduction}\label{sec:Intro}

Probabilistic spatio-temporal (ST) forecasting is integral to several disciplines in the environmental sciences such as ecology, meteorology, and oceanography. Often, such forecasts are constructed using statistical ST models, which can be broadly grouped into two categories: marginal (or geostatistical) ST models, and dynamic ST models (DSTMs). The former models are built using ST covariance functions, which encode the marginal dependencies between the variable of interest at two different locations in space and time \citep[e.g.,][]{Cressie_1999, Gneiting_2007b, montero2015spatial}. The latter models (in particular, the discrete-time variants) are generally constructed from conditional dependence relationships between two spatial fields at two consecutive time steps  \citep[see the overviews in][Chapter 5]{Cressie_2011, Wikle_2019}.  Marginal models tend to provide simple descriptions of the underlying phenomena, they are general-purpose, and can be easily implemented using readily-available software packages. DSTMs tend to be application-specific, however they can encode \emph{mechanistic descriptions} of the processes being modelled, which makes them well-suited for forecasting purposes \citep[e.g.,][Chapters 5 and 6]{Wikle_2019}.

One of the biggest challenges with statistical implementations of DSTMs is the specification of realistic structure (e.g., nonlinearity) in a manner that is parsimonious and that can accommodate uncertainty quantification (UQ).  There are some parametric DSTMs that have been developed to accommodate complex structure, notably those in the class of general quadratic nonlinear models \citep{wikle2010general}, which are flexible and which have been shown to be suitable for modelling many real-world processes. Since these models are highly parameterised, one typically employs process dimension reduction, and/or regularisation when making inference, either by directly incorporating knowledge about the underlying dynamics of the system of interest  and/or through prior specification within a multi-level (deep) Bayesian hierarchical modelling (BHM) framework \citep[e.g.,][]{wikle2001spatiotemporal, wikle2010general,  leeds2014emulator}. Quadratic nonlinear models tend to be quite complex, they require a relatively large amount of training data, and are computationally challenging to implement.
 
The BHM statistical approach commonly employed for statistical DSTMs builds dependencies in complex processes through the marginalisation of a series of conditional models.  Similarly, the deep neural network models in machine learning (ML) that have seen great success in image and language processing (e.g., through convolutional neural networks (CNNs) or recurrent neural networks (RNNs)) are also based on a sequence of linked models (typically, deterministic models), where the outputs from one level are the inputs to the next level \citep[see][for an overview]{goodfellow2016deep}. As described in \cite{wikle2019comparison}, the deep ML framework has many features in common with the BHM approach in the context of ST models. For example, both approaches consider highly parameterised models, both require large amounts of training data, and both require some form of regularisation.  There are fundamental differences as well, primarily that, on the contrary to conventional statistical models, the deep ML models are geared to provide the best possible out-of-sample prediction performance, and are less equipped to answer questions concerning UQ and model interpretability.  Thus, there is a unique opportunity to place deep ML models into a statistical framework so as to harness their potential in a more formal inference setting.

The similarities and differences discussed above have helped establish a new branch in Statistical Science that looks at combining formal statistical models with the flexibility of deep ML models with a view to exploiting the strengths of both approaches in a single framework for better prediction and forecasting. 
For example, \citet{Nguyen_2019} use a type of RNN known as the long short-term memory (LSTM) within a classic stochastic volatility model in order to cater for long-range (temporal) dependence, while \citet{Tran_2019} employ neural networks within the classic generalised linear mixed model framework. In the context of ST statistics, \cite{mcdermott2017ensemble}  consider a simple ensemble (parametric bootstrap) forecasting approach with a type of RNN known as an echo state network (ESN), while \cite{mcdermott2019deep} use deep ESN models as basis function generators that are then used within a generalised additive mixed model. Although the ESN approaches have shown remarkable success, they suffer from an interpretability point of view. Recently, \cite{Zammit_2019} used a deep structure to model nonstationarity in spatial models. The resulting model is interpretable, but the framework is firmly seated within the standard geostatistical setting where time, if considered, would be treated as an extra dimension; such models tend to be ill-suited for forecasting purposes. To the best of our knowledge there is no work that explores combining the flexibility of deep ML models with the interpretable and inferential advantages of statistical DSTMs. Here we provide a first step to remedying this by presenting a novel approach whereby we use a CNN to encode nonlinear dynamics or temporally-varying dynamics in a statistical integro-difference equation (IDE) model to facilitate probabilistic forecasting.

IDE models have proven to be very suitable for modelling dynamical processes \citep[e.g.,][]{Kot_1986,Kot_1996,Wikle_1999,Wikle_2002,Xu_2005,Calder_2011,Zammit_2012}. These models describe the conditional dependence between the spatial field at a future time point and the field at the present-time point through an integral operator.  Typically the operator is assumed to be linear, but this assumption is rarely tenable in practice over large time horizons. Nonlinearity has been addressed in quadratic nonlinear IDE DSTMs \citep[e.g.,][]{wikle2011polynomial}, but these IDEs typically require a reduced state dimension and can be computationally very difficult to work with, as is also often the case with other quadratic nonlinear DSTMs.

In this article we address the issue of nonlinearity in an IDE by using a CNN to learn about parameters governing the dynamics from the most recent behaviour of the (partially observed) process. The framework builds on that of \citet{deBezenac_2018} who considered a purely deterministic setting where the process is completely observed, and involves introducing state dependence into the operator of the statistical IDE. The CNN is fitted offline to extract process dynamics from the most recent process behaviour, but once it is fitted, probabilistic forecasting is implemented extremely quickly online using an ensemble Kalman filter with no requirement for repeated parameter estimation. Further, we show that the CNN provides a global prior model for the dynamics that is realistic and interpretable. Indeed, we show that the learned representation is so powerful that the CNN-IDE trained in one application context can be used for successfully producing probabilistic forecasts in an entirely different application context where the dynamics are partially governed by the same underlying physical principles (e.g., advection and diffusion). The resulting framework achieves the desired aim of harnessing the model flexibility inherent to CNNs and the interpretative and probabilistic prediction advantages of statistical DSTMs. 

The article is organised as follows. In Section 2 we describe the statistical IDE model and provide simplifications to \citet{deBezenac_2018}'s model based on rank-reduction methods. In Section 3 we then place the CNN-IDE into a hierarchical statistical ST modelling framework, and implement an ensemble Kalman filter from which we are able to obtain filtered predictions and forecasts of both the process and the process dynamics. In Section 4 we use the CNN-IDE for providing one day forecasts of sea-surface temperatures in the North Atlantic Ocean and 10-minute nowcasts of radar reflectivities in Sydney. Section 5 concludes through a discussion and an outline of future extensions.

\section{Modelling and inferential framework}

In Section~\ref{sec:IDE-DSTM} we give a brief account of the IDE model while in Section~\ref{sec:IDE-DSTM2} we cast the IDE into a state-dependent model by expressing the operator parameters as a function of the process at current and preceding time steps. In Section~\ref{sec:CNN} we justify the use of CNNs for representing the mapping between the process and the parameters governing the system dynamics. 

\subsection{Background to the IDE-DSTM}\label{sec:IDE-DSTM}

The IDE DSTM finds its origins in models that describe the advection or spread of spatially-referenced variables in time.  Let $Y_t(\cdot)$ denote a spatial process on some domain $D$ at time $t$. The vanilla first-order linear IDE \citep[e.g.,][]{Wikle_1999} is given by
\begin{equation}\label{eq:IDE}
Y_{t+1}(\svec) = \int_D m(\svec, \uvec; \thetab_t(\svec))Y_{t}(\uvec)\intd \uvec + \eta_t(\svec); \quad \svec \in D, 
\end{equation}
where $\eta_t(\cdot)$ is some spatially correlated disturbance or random forcing on $D$ that accounts for model discrepancy, and $m(\cdot, \cdot\,;\thetab_t(\cdot))$ is a mixing kernel parameterised by a spatio-temporally varying parameter vector $\thetab_t(\cdot)$.

Several works show that the IDE is a physically-interpretable statistical DSTM with good predictive ability. Variants of \eqref{eq:IDE} have been used, for example, to model the spread (dispersion) of invading organisms \citep{Kot_1986, Kot_1996}, cloud data \citep{Wikle_2002}, the spread and movement of storm cells \citep[]{Xu_2005}, aerosol optical depth \citep{Calder_2011}, electroencephalogram signals \citep{Freestone_2011}, and conflict events \citep{Zammit_2012}.

The mixing kernel is the most important component of the IDE as it  governs the dynamics of the modelled ST process. For example, if $m(\svec, \uvec; \thetab_t(\svec)) \equiv m_0(\|\svec - \uvec\|/d)$ for  $d > 0$ and $t= 1,\dots,$ then there is no advection, or drift, being modelled. If, moreover, $m_0(\cdot)$ is the squared exponential kernel and $\eta_t(\cdot) = 0$, then $Y_{t+1}(\cdot)$ is the solution to the heat equation with $Y_t(\cdot)$ as initial condition \citep[e.g.,][]{Coleman_2005}. In this context, $d$ is a diffusion parameter and, if made spatially varying, can be used to describe spatially-varying diffusion. Advection, or drift, can be modelled by shifting the kernel from the origin. For example, if $m(\svec, \uvec; \thetab_t(\svec))  \equiv m_0(\|\svec - \vvec - \uvec\|/d)$, then $\vvec$ takes on the role of advection parameters.  Even so, the modelled dynamics will be the same everywhere in $D$ unless $\vvec$ is made to vary spatially. Indeed, complex spatially-varying dynamics can only be captured if both $d$ and $\vvec$ are functions of space at any given point in time. Letting the spatially-varying kernel parameters vary in time yields the general model \eqref{eq:IDE}.

The requirement to make $\thetab_t(\cdot)$ time-varying often stems from the physical behaviour of many ST processes of interest: these processes tend to be highly nonlinear, but local linearity (in time) is often a reasonable assumption.  Treating the kernel parameters as time-varying to cater for nonlinearity, however, comes with the downside that these parameters need to be estimated for each $t$. Further, the estimated parameters only impart information on the system dynamics at specific points in time. While such models perform remarkably well when the nonlinearity is mild or the true dynamics do indeed vary slowly in time (see Section~\ref{sec:application}), ideally we have at our disposal the nonlinear mapping itself. Knowledge of this mapping would increase our understanding of the process' \emph{global} (as opposed to \emph{local}) behaviour, and relieve the analyst from having to estimate the parameters for each time $t$. There have been attempts elsewhere to find the nonlinear operator in the IDE directly; for example \citet{wikle2011polynomial} cast the IDE into a polynomial nonlinear framework that in turn can be cast as a state-dependent model. However, the requirement for dimension reduction and the computational difficulties often encountered with these models have hindered  their widespread practical use.

In the next section we propose modelling the process' global behaviour by recasting the IDE into a state-dependent model where the parameters $\thetab_t(\cdot)$ are formulated as functions of $\{Y_{t'}(\cdot): t' \le t\}$ using CNNs. This reformulation yields a state-dependent mixing kernel with a deep learning model encoding a complex mapping, but ultimately a statistical model that is a member of the general quadratic nonlinear family of models. As we show in Section~\ref{sec:application}, the resulting model is extremely amenable to describing ST evolving dynamics of ST phenomena, so much so that a fitted model can be used for forecasting in other environmental applications that exhibit similar dynamical behaviour to that in which the IDE was originally fitted (without a need to re-estimate any parameters describing the dynamics).

\subsection{The IDE-DSTM with state-dependent kernel}\label{sec:IDE-DSTM2}

Our starting point is the framework of \citet{deBezenac_2018}, who took a radically different view of the \emph{deterministic} IDE model. Instead of establishing a parametric model for $m(\cdot, \cdot\,; \thetab_t(\cdot))$, they instead propose finding a mapping between $\thetab(\cdot)$ and the set $\Yset^{(\tau)}_t(\cdot) \equiv \{Y_t(\cdot),\dots,Y_{t-\tau+1}(\cdot)\}$, where $\tau \ge 2$. Once this mapping is found, $\Yset^{(\tau)}_t(\cdot)$ is used to determine the mixing kernel and hence predict $Y_{t+1}(\cdot)$, which is then used to evaluate the mixing kernel at time $t+1$.
 
This approach to modelling the mixing kernel is based on the assumption that the spatially-varying dynamics of the process at time $t$ are determined in some nonlinear fashion by the process' behaviour at the most recent $\tau$ time instants, where $\tau$ is pre-specified. \citet{deBezenac_2018} manage to find a good approximation to this mapping using CNNs, inspired by their ubiquitous use in characterising the motion of objects between two images  \citep[in solving what is known as the optical flow problem; see][]{Dosovitskiy_2015}. Their results show that it is indeed possible to learn this complex mapping given sufficient (in their case, several tens of thousands) sequences of images. 
However, they considered the purely deterministic context (with no UQ) where the images are completely observed, and made the implicit assumption that $\eta_t(\cdot)$ is spatially uncorrelated which, as we show in our application study, is an untenable assumption.

An alternative modelling strategy, which we explore next, is to place the resulting CNN mapping within the statistical IDE framework \eqref{eq:IDE}. In our framework, \cite{deBezenac_2018}'s formulation translates to a \emph{state-dependent} mixing kernel $k(\cdot, \cdot\,; \thetab(\cdot\,;\,\Yset^{(\tau)}_t,\psib))$ where $\psib$ are some unknown parameters determining the mapping between $\Yset^{(\tau)}_t(\cdot)$ and  the spatially-varying parameters.   That is, this function takes the process' most recent behaviour and translates it to mixing kernel parameters.  Crucially, $\thetab(\cdot)$ is now a \emph{time-invariant function} in the sense that, although its output varies in time, the {\it functional relationship} between the most recent values of the process and the mixing kernel parameters does not vary in time (see details below).   Importantly, all of the modelling effort is then placed on finding the (extremely complex) nonlinear relationship between $\Yset^{(\tau)}_t(\cdot)$ and $\thetab(\cdot)$ through some model parameterised by $\psib$. This framework allows for spatially-correlated model mismatches and, compellingly, since within the hierarchical framework $Y_t(\cdot)$ is a stochastic process, the model has a  state-dependent kernel that is itself random. In this way, uncertainty on the process leads to uncertainty on the dynamics, which leads to uncertainty in the predictions, and which in turn could be useful for UQ.

The state-dependent IDE is given by
\begin{equation}\label{eq:IDE2}
Y_{t+1}(\svec) = \int_D k(\svec, \uvec; \thetab(\svec; \Yset_t^{(\tau)}, \psib))Y_{t}(\uvec)\intd \uvec + \eta_t(\svec); \quad \svec \in D, 
\end{equation}
where $\eta_t(\cdot)$ is a zero-mean spatial Gaussian process with covariance function $C(\cdot, \cdot\, ; \alphab)$, and $\alphab$ are unknown parameters. In this work we use a squared-exponential kernel, which we define as
\begin{equation}\label{eq:kernel}
k(\svec, \uvec; \thetab(\svec; \Yset_t^{(\tau)}, \psib)) \equiv \frac{1}{4\pi \theta_{1}(\svec; \Yset_t^{(\tau)}, \psib)}\exp\left(-\frac{h(\svec, \uvec; \thetab(\svec; \Yset_t^{(\tau)}, \psib))^2}{4\theta_{1}(\svec; \Yset_t^{(\tau)}, \psib)}\right),
\end{equation}
where
$$
h(\svec, \uvec; \thetab(\svec; \Yset_t^{(\tau)}, \psib)) \equiv \left\| \svec - \begin{pmatrix} \theta_{2}(\svec; \Yset_t^{(\tau)}, \psib) \\ \theta_{3}(\svec; \Yset_t^{(\tau)}, \psib)\end{pmatrix} - \uvec \right\|.
  $$
As discussed in Section~\ref{sec:IDE-DSTM}, the state-dependent spatially-varying parameters $\thetab(\cdot)$ have a physical interpretation; specifically $\theta_{1}(\cdot)$ describes diffusivity and $(\theta_{2}(\cdot), \theta_{3}(\cdot))'$ describe process advection. Equation~\ref{eq:kernel} is similar to that of \citet{deBezenac_2018} who used a full-rank model for $\thetab(\cdot)$ and assumed that $\theta_{1}(\cdot)$ was a fixed, known constant that does not vary with space or time. We choose the squared-exponential kernel because it is interpretable and because it is ubiquitously used in IDE models. We note, however, that other more sophisticated kernels may be readily used within this framework if needed \citep{Richardson_2017, Richardson_2018}. 

It is typically reasonable to assume that the spatially-varying dynamics vary smoothly in space. Here, therefore, we further decompose each component of $\thetab(\cdot)$ using a sum of $r$ radial basis functions $\{\phi_j\}$ to yield
\begin{equation}\label{eq:theta}
\theta_{i}(\cdot\,;  \Yset_t^{(\tau)}, \psib) = \sum_{j = 1}^r \phi_j(\cdot)\omega_{ij}(\Yset_t^{(\tau)}; \psib); \quad i = 1,2,3,
\end{equation}
where $\{\omega_{ij}\}$ are basis-function weights that are state-dependent. 

Consider a regular fine discretisation of our domain $D$, $D^G$. Let $\Yvec_t$ and $\etab_t$ be $Y_t(\cdot)$ and $\eta_t(\cdot)$ evaluated on $D^G$, respectively, and define $\Yvec_t^{(\tau)} \equiv (\Yvec_t', \dots,\Yvec_{t-\tau+1}')'$. An analogue to \eqref{eq:theta} based on this discretisation is
\begin{equation}\label{eq:theta2}
\vartheta_{i}(\cdot\,;  \Yvec_t^{(\tau)}, \psib) = \sum_{j = 1}^r \phi_j(\cdot)w_{ij}(\Yvec_t^{(\tau)}; \psib); \quad i = 1,2,3,
\end{equation}
where now $\vartheta_i(\cdot)$ and $w_{ij}(\cdot)$ are identical to $\theta_i(\cdot)$ and $\omega_{ij}(\cdot)$, respectively, except that they take discretised processes as inputs.

The IDE model in \eqref{eq:IDE2} is defined on the lattice as follows,
\begin{equation}\label{eq:processmodel}
  \Yvec_{t+1} = \Kmat(\Yvec^{(\tau)}_t; \psib)\Yvec_t + \etab_t.
\end{equation}
 The matrix $\Kmat(\cdot)$ is constructed by replacing $\thetab(\svec; \Yset_t^{(t)}, \psib)$ with $\varthetab(\svec; \Yvec_t^{(t)}, \psib)$ in \eqref{eq:kernel}, evaluating the resulting expression over $D^G \times D^G$, and multiplying the elements by the area of a single grid cell in the discretisation, thereby approximating the integral in \eqref{eq:IDE2} as a Riemann sum. The process model given in \eqref{eq:processmodel} is clearly state-dependent in the sense that the transition matrix $\Kmat(\cdot)$ depends on $\Yvec_t$. This form of interaction is also quadratic in nature, and thus this process model sits firmly within the quadratic nonlinear DSTM class of \cite{wikle2010general}.

Assume now that we have a time series of the discretised process $\Yvec_1,\dots, \Yvec_T$ (these would often come in the form of an image sequence). The discrete-space IDE model is a multivariate, $\tau$-order, Markov model. The conditional likelihood function (conditional on the first $\tau$ images in the series) of the unknown parameters $\{\psib, \alphab\}$ is
\begin{equation}\label{eq:Lik}
  L(\psib, \alphab) 
  = \prod_{t = \tau}^{T-1}p(\Yvec_{t+1} \mid \Yvec_{t}^{(\tau)}, \psib, \alphab) \equiv \prod_{t = \tau}^{T - 1}L_t(\psib, \alphab),
\end{equation}
where $L_t(\psib, \alphab) \equiv p(\Yvec_{t+1} \mid \Yvec_{t}^{(\tau)}, \psib, \alphab)$. For very large $T$, \eqref{eq:Lik} generally yields maximum likelihood estimates that are practically identical to those when considering the marginal likelihood function $p(\Yvec_{T}^{(T)} \mid \psib, \alphab)$. Importantly, we have that 
\begin{equation}\label{eq:Ycond}
\Yvec_{t+1}  \mid \Yvec_t^{(\tau)},\psib, \alphab \sim \Gau(\Kmat(\Yvec^{(\tau)}_t; \psib)\Yvec_t, \Sigmamat_\alphab),
\end{equation}
where $\Sigmamat_\alphab \equiv \var(\etab_t),\, t= \tau,\dots, (T-1)$. For some random subset $\Tset$ of $\{\tau,\dots,(T-1)\}$, we therefore have that $\frac{T - \tau}{|\Tset|}\sum_{t\in\Tset} \log L_t(\psib,\alphab)$ is an unbiased estimator of $\log L(\psib, \alphab)$ and $ \frac{T - \tau}{|\Tset|}(\sum_{t\in\Tset}\nablab\log L_t(\psib,\alphab))$ is an unbiased estimator of $\nablab \log L(\psib, \alphab)$ \citep[see, e.g., ][Appendix B, for details]{Zammit_2019}. We can therefore use stochastic gradient ascent for maximising the conditional log-likelihood function, where at each step in the algorithm we use $|\Tset|$  randomly selected sets of $\{\Yvec_{t+1}, \Yvec_t^{(\tau)}\}$ (also known as minibatches). This approach allows us to obtain maximum (conditional) likelihood estimates in a computationally-efficient manner.

\subsection{Using CNNs to encode spatio-temporal dependency}\label{sec:CNN}

Using the process on the discrete lattice and evaluating \eqref{eq:theta2} on $D^G$, we obtain the parameter vectors $\varthetab_{i}(\Yvec_t^{(\tau)}; \psib) = \Phimat\wvec_{i}(\Yvec_t^{(\tau)}; \psib)$, where $\Phimat \equiv (\phib(\svec): \svec \in D^G)'$ and $\wvec_i(\Yvec_t^{(\tau)}; \psib) \equiv (w_{i1}(\Yvec_t^{(\tau)}; \psib),\dots,w_{ir}(\Yvec_t^{(\tau)}; \psib))', i = 1,2,3$.  The time-invariant functional relationship between $\Yvec_t^{(\tau)}$ and $\wvec_{i}$ is not straightforward. However, it is plausible that the relationship between \emph{features} of $\Yvec_t^{(\tau)}$ and  $\wvec_{i}$ is straightforward. For example, a horizontal positive shift of mass in the process across three consecutive time steps in a certain region is representative of a large positive weight $w_{2j^*}$, where $j^*$ indexes a radial basis function located within that region. Conversely, a horizontal negative shift is representative of a large negative weight, while no shift is representative of a zero weight. 

\begin{figure}[t!]
  \includegraphics[width = \textwidth]{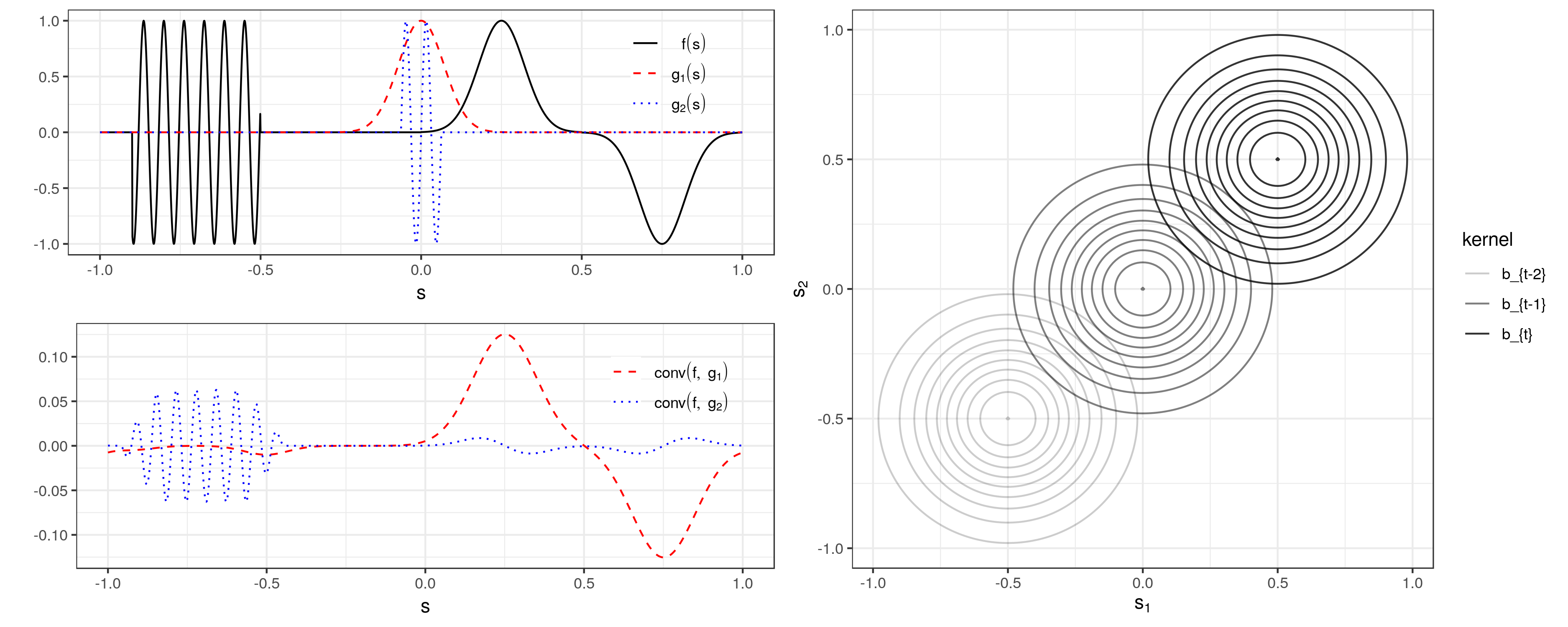}
  \caption{Left: Toy example illustrating signal convolution. In the top panel, the black solid line denotes the signal while the red dashed and blue dotted lines denote two different convolution kernels. In the bottom panel, the red dashed and blue dotted lines show the result of convolving the signal in the top panel with the convolution kernels denoted by the red dashed and blue dotted lines in the top panel, respectively. Note how the convolution operation filters out signal features that are distinct from the kernel. Right: A sketch of a 3D convolution kernel (or 3D filter) used to extract a north-easterly direction of motion from three spatial images ordered in time.}\label{fig:Convexample}
\end{figure}

The problem therefore reduces to extracting such features from a sequence of process realisations. In signal processing, signal detection is often done using convolutions. For illustration, let us return to the continuous case and consider a one-dimensional function, $f(\cdot)$, to represent a signal, and another one-dimensional function of compact support, $g(\cdot)$ (which we also call a \emph{filter}),  to encode a feature. The convolution of $f(\cdot)$ and $g(\cdot)$ returns a function with an absolute value that is large in regions where $f(\cdot)$ exhibits a feature similar to that encoded in $g(\cdot)$, and an absolute value that is small otherwise; see Fig.~\ref{fig:Convexample}, left panel, for an example. Now, the stack of spatial processes in $\Yset_t^{(\tau)}(\cdot)$ is three-dimensional, with the third dimension denoting lag. In this case we can carry out feature extraction using three-dimensional filters, specifically by summing up, point-wise, the output from 2D convolutions done for each lag. In this way, the filter extracts a dynamic feature of the process across time; see Fig.~\ref{fig:Convexample}, right panel, for an example of such a filter. Feature extraction of this nature is precisely what CNNs were designed for.

\begin{figure}
  \includegraphics[width = \textwidth]{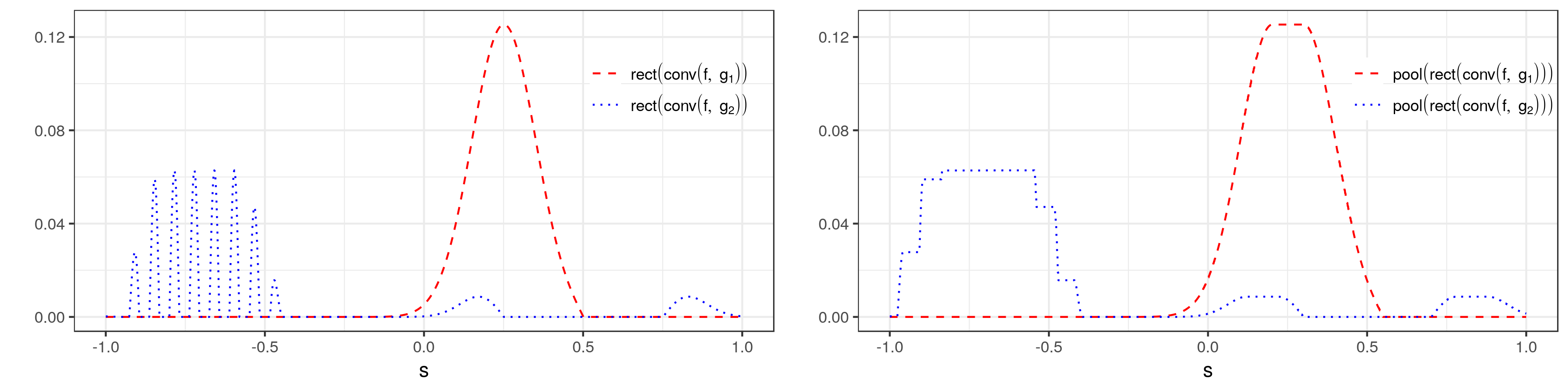}
  \caption{Left: Output obtained when taking the convolved signals in Fig.~\ref{fig:Convexample} and keeping only the positive components, in an action known as rectification. Right: Output obtained when taking the rectified signals in the left panel and max-pooling them over subregions of width $0.05$. \label{fig:rect_pool}}
\end{figure}

Consider again our spatially-discretised IDE, and denote the output of the $k$th set of 2D convolutions on $\Yvec_t^{(\tau)}$ as $\Fmat^{(1)}_{t,k}$. Let $Y_t[i,j]$ denote the $(i,j)$th pixel in the image represented by $\Yvec_t$. We have that
\begin{equation}\label{eq:discreteconv}
F^{(1)}_{t,k}[i,j] = \sum_{q = 0}^{\tau - 1}\left(\sum_{l,m}Y_{t - q}[i - l, j - m]g^{(1)}_{q,k}[l,m]\right),
\end{equation}
where the limits of the inner sum in \eqref{eq:discreteconv} depend on the two-dimensional spatial support (sometimes referred to as the patch size) of the discretised three-dimensional filter given by  $\{\gvec^{(1)}_{0,k},\dots,\gvec^{(1)}_{\tau-1,k}\}$.

Typically, in CNNs, detection of a feature at the $(i,j)$th pixel is done by passing each value in $\Fmat^{(1)}_{t,k}$ through a rectified linear unit, which simply returns $F^{(1)}_{t,k}[i,j]$ if $F^{(1)}_{t,k}[i,j] > 0$ and zero otherwise. Usually, the output image is also smoothed out, by either taking a moving-average (average-pooling) or a moving-maximum (max-pooling) of the rectified $\Fmat_{t,k}^{(1)}$, and then subsampled (in what is known as a stride). Following this `post-processing' of $\Fmat_{t,k}^{(1)}$, one ends up with $\tilde\Fmat_{t,k}^{(1)}$ which contains a (lower-resolution) image encoding the locations and strengths of a certain dynamic (encoded in $\{\gvec^{(1)}_{0,k},\dots,\gvec^{(1)}_{\tau-1,k}\}$) in $\Yvec_t^{(\tau)}$. In Fig.~\ref{fig:rect_pool} we illustrate the action of rectifying and  max-pooling when doing feature extraction in our one-dimensional example. Note how the final convolved, rectified, and pooled output reflects where in the input domain the features encoded in the convolution kernels are apparent in the signal.

One could stop here and then fit a linear mapping from $\tilde\Fmat_{t,k}^{(1)}, k = 1,2,\dots,\,$ to $\wvec_{i}$. This map, however, may still be relatively complex. In conventional CNNs one applies convolution operations to $\tilde\Fmat_{t,k}^{(1)}$ (followed by rectifying, pooling and subsampling) until the dimension of the images at the other end of the network, $\tilde\Fmat_{t,k}^{(n)}$ say, is of a similar dimension as that of the output, in our case $\wvec_{i}$. The relationship between $\tilde\Fmat_{t,k}^{(n)}$ and $\wvec_{i}$ is then modelled using a linear map. Convolutions are therefore repeatedly used to `drill down' information in a set of images $\{\Yvec_{t+1}, \Yvec_t^{(\tau)}\}$ into features that can be used to easily model $\wvec_{i}$.

The parameters in the CNN, $\psib$, are those that define the filters at each stage. The number of unknown parameters can be large. For example, if $\tau = 3$, the patch size of $\gvec_{r,k}^{(1)}$ is $5 \times 5$, and 64 filters are used in the first layer, then just for the first stage there are 4800 parameters that need to be estimated. It is not uncommon for such models to have tens to hundreds of thousands of parameters that need to be estimated using maximum likelihood; regularisation techniques such as dropout are often needed to avoid overfitting \citep{Srivastava_2014}.

\begin{figure}[t!]
  \begin{center}
    \includegraphics[width=0.8\textwidth]{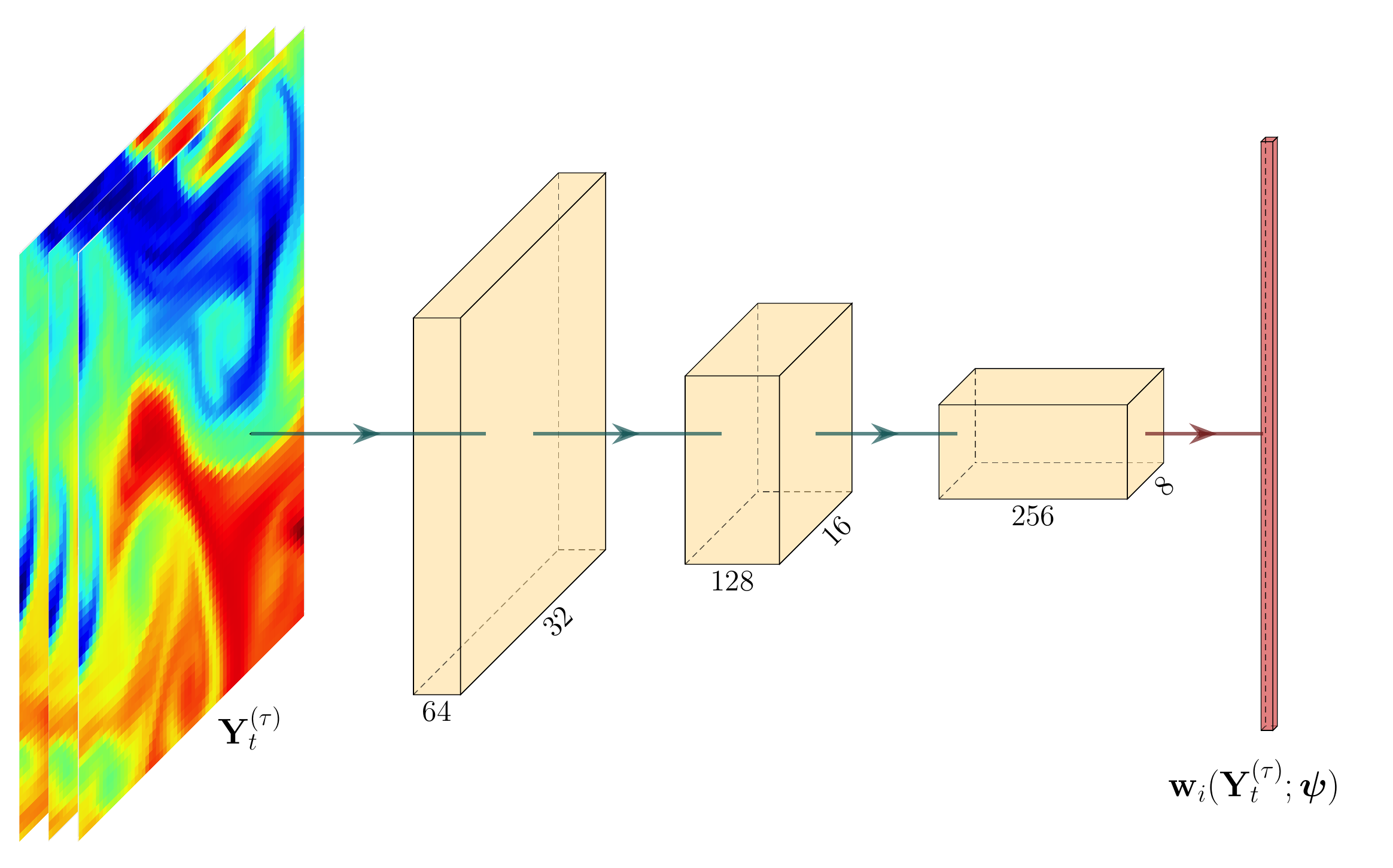}
   \caption{Sketch of CNN architecture used in the CNN-IDE to construct the map $\Yvec_t^{(\tau)} \xrightarrow[]{ ~\psib~} \wvec_{i}, i = 1,2,3$. The $\tau = 3$ temporally ordered input spatial images of size $64 \times 64$ are convolved with 64 3D filters, rectified, max-pooled, and subsampled to yield 64 spatial images of size $32 \times 32$. The process is repeated with 128 and 256 3D filters, respectively, to yield 256 8$\times$8 images. These 256 images are then vectorised and premultiplied by a matrix $\Amat_i$ to yield the vector $\wvec_i$.\label{fig:CNNexample}}
  \end{center} 
\end{figure}

In our framework, the dimension of $\wvec_{i}$ determines how many layers to use within the CNN. In Fig.~\ref{fig:CNNexample} we illustrate the architecture we use in the simulation and application study for when $\tau = 3$, which follows closely the first part of that used by \citet{deBezenac_2018}. We apply 64 filters to the set of three input images in $\Yvec^{(3)}_t$ of size $64 \times 64$ each. These convolutions are followed by a rectified linear unit, and a max-pooling unit, to yield $\tilde\Fmat^{(1)}_{t,k}, k = 1,\dots, 64$, where each $\tilde\Fmat^{(1)}_{t,k}$ represents an image of size $32 \times 32$. We repeat the process with 128 filters to obtain $\tilde\Fmat^{(2)}_{t,k}, k = 1,\dots, 128$, representing images of size $16 \times 16$, and with 256 filters to obtain $\tilde\Fmat^{(3)}_{t,k}, k = 1,\dots, 256$, representing images of size $8 \times 8$. We use a simple linear model in the final layer, that is, we let $\wvec_{i} = \Amat_i\overline\Fmat^{(3)}_{t},i = 1,\dots, 3$, where $\overline\Fmat^{(3)}_{t} \equiv (\tilde\Fmat^{(3)'}_{t,k}: k = 1,\dots, 256)'$, and $\Amat_i$ is an unstructured weights matrix that is also estimated.

It is easiest to use separate CNNs (with identical architectures) to model each of the three kernel parameters. We can, however, take advantage of the fact that filters that yield important features in the horizontal direction are simply the transpose of filters that yield important features in the vertical direction. By constraining the first-layer filters associated with $\wvec_{2}$ to be the transpose of those associated with $\wvec_{3}$, and by subsequently setting all other filters to be the same and $\Amat_2 = \Amat_3$, we reduce the amount of parameters in the model by one-third.

\section{Inference with the CNN-IDE}\label{sec:inference}

In Section~\ref{sec:EnKF} we place the IDE with the CNN-driven kernel inside a hierarchical structure where we separate the data model from the process model; this allows us to consider the case where $\{Y_t(\cdot)\}$ is not fully observed, and to use the model in a data assimilation setting. In Section~\ref{sec:guidelines} we provide some practical guidelines for implementing the CNN-IDE.

\subsection{Approximate filtering and forecasting}\label{sec:EnKF}

Assume that we have a set of irregularly-spaced point-referenced measurements $\Zvec_t, t = \tau+1,\dots,T$, where $\Zvec_t \equiv (Z_{t,1},\dots,Z_{t,m_t})'$. These data could be, for example, in-situ readings of carbon-dioxide concentration, or remote sensing retrievals of sea-surface temperature (SST). We model these data as
\begin{equation}\label{eq:datamodel}
\Zvec_t = \Hmat_t\Yvec_t + \epsilonb_t,\quad t = \tau+1,\dots,T,
\end{equation}
where $\Yvec_t$ is $Y_t(\cdot)$ evaluated over $D^G$, $\Hmat_t$ is an incidence matrix identifying which pixels the measurements are in \citep[e.g.,][Chapter 5]{Wikle_2019}, and $\epsilonb_t$ is Gaussian measurement error. The data model \eqref{eq:datamodel} combined with the process model \eqref{eq:processmodel} yields a high-dimensional $\tau$-order state-space model.

It is straightforward to see that the $\tau$-order model for $\Yvec_t$ is also a first-order process model for $\Yvec_t^{(\tau)}$ \citep[e.g.,][Section 13.1]{Hamilton_1994}. Specifically,
\begin{equation}
\Yvec_{t+1}^{(\tau)} = \Kmat^{(\tau)}(\Yvec_t^{(\tau)};\psib)\Yvec_{t}^{(\tau)} + \etab_t^{(\tau)},
\end{equation}
where
$$
\Kmat^{(\tau)}(\Yvec_t^{(\tau)};\psib) \equiv
\begin{pmatrix}
  \Kmat(\Yvec_t^{(\tau)};\psib) & \zerob & \zerob & \cdots & \zerob \\
  \Imat                           & \zerob & \zerob & \cdots & \zerob \\
  \zerob                          & \Imat  & \zerob & \cdots & \zerob \\
  \vdots                          & \vdots & \vdots & \ddots & \vdots \\
  \zerob                          & \zerob & \zerob & \cdots & \Imat
\end{pmatrix},
$$
and $\etab_t^{(\tau)} \equiv (\etab_t', \zerob', \hdots, \zerob')'$. This is a useful representation, as sequential inferential algorithms are readily available for first-order state-space models. Since the dimension of $\Yvec_t^{(\tau)}$ is large and, more importantly, the transition matrix is state-dependent, one can employ the ensemble Kalman filter (EnKF), or variants thereof, to predict, forecast, or hindcast $\Yvec_t^{(\tau)}$ from available data $\{\Zvec_t\}$. Algorithmic and implementation details for the EnKF are available from several sources; see \citet{Katzfuss_2016b} for a recent review.

There are several CNN and variance parameters that need to be estimated in this framework. 
Parameter estimation with the EnKF can generally be done in an iterative framework  \citep[e.g.,][]{Gibson_2005, Zammit_2011, Katzfuss_2019}; since such algorithms are well-established, we omit details. An advantage of the CNN-IDE, however, is that parameters corresponding to the CNN can be reasonably estimated offline using complete data generated from a numerical physical model or an analyses (this is the strategy we adopt in Section~\ref{sec:application}). Once this is done, unlike in conventional DSTMs, the parameters corresponding to the dynamics do not need to be estimated online as these are implicitly encoded in the process' most recent behaviour. The implication of this is huge from a computational standpoint: If one also has reliable estimates of the other (typically variance) parameters within the model, the CNN-IDE could be used for practically-effective  online prediction without doing any parameter estimation!

A further interesting consequence of using an EnKF in conjunction with a state-dependent transition matrix is that uncertainty in the process dynamics at time $t$ is induced by uncertainty in $\Yvec_t^{(\tau)}$. In particular, recall that the flow vectors and diffusion basis-function coefficients are given by $\wvec_{i}(\Yvec_t^{(\tau)}; \psib)$. For the $j$th ensemble member $\Yvec_t^{(\tau), j}$ we have a corresponding vector of coefficients $\wvec_{i}(\Yvec_t^{(\tau), j}; \psib)$. From a collection of $N$ ensemble members $\{\Yvec_{t}^{(\tau), j}: j = 1,\dots, N \}$ we can therefore calculate the empirical mean and variance of these dynamical basis-function coefficients. These quantities would constitute \emph{filtered dynamics} if the ensemble members are treated as samples from the process conditioned on $\Zvec_{1:t}$; an attractive feature is that one can also obtain \emph{forecasted dynamics} if the ensemble members are forecasts, that is, are conditioned on $\Zvec_{1:t'}$ where $t' < t$. While uncertainty in the forecasts of the dynamics can be expected to grow quite rapidly, this capability is novel, and may be useful when the dynamics tend to vary slowly in time. A fortuitous benefit of this model is that these uncertainties over the dynamical parameters are obtained for free, without the need for further computations.  This is a considerable advantage over the use of conventional Bayesian hierarchical models where uncertainty in the dynamics are generally obtained via computationally intensive Markov chain Monte Carlo techniques \citep[e.g.,][]{Wikle_2002} or bootstrap.

\subsection{Implementation considerations}\label{sec:guidelines}

In the preceding sections we discussed the general modelling and inferential framework behind the CNN-IDE. Here, we list three issues that require consideration when implementing the CNN-IDE in practice.

\begin{itemize}
\item \emph{Computation}: Parameter estimation of $\psib$ in the CNN component of the CNN-IDE needs to be done on graphical processing units (GPUs), which are able to carry out the required linear algebraic operations needed to compute the log-likelihood (and its gradient) corresponding to a minibatch in parallel. Somewhat serendipitously,  the same code that is used to compute the predictions for the likelihood (via the CNN) for a minibatch in parallel, is exactly the same code that is needed for computing the predictions of the ensemble members in the EnKF. At each time step, the ensemble members are communicated back to the main processing unit, which then feeds them back to the GPU as inputs for the next time step. 

\item \emph{Boundary effects:}  For simplicity, we have let the integration in \eqref{eq:IDE2} be over $D$, as is conventional in IDEs. However, $D$ is more often than not a subregion of interest, embedded within a larger region in which the ST process is evolving. This model can therefore be a poor representation of reality at the boundaries. A crude way for dealing with the boundary effects, which appears to work well in our application of Section~\ref{sec:application}, is to subdivide $D$ into an interior region $D^{\intr}$ and an exterior region $D^{\ext}$, and to only make inference on the process inside $D^{\intr}$. In practice this means constructing $D$ such that the boundary is well-buffered from the region of interest $D^{\intr}$.

\item \emph{EnKF localisation:} Sample covariance matrices, calculated when doing EnKF updates, often include spurious covariances due to the relatively small number of ensemble members used. This is also true in our application: If $\tau = 3$ and the images we use are of size $64 \times 64$, then each ensemble member is 12288-dimensional, whereas we might be using on the order of 64 ensemble members in a typical EnKF framework. Localisation is the process by which the sample covariance matrices are \emph{tapered} \citep{Furrer_2006}, usually using inter-pixel distance as criterion for tapering, to remove these spurious covariances. Localisation was essential for providing realistic predictions and forecasts in our application study.

\end{itemize}

\section{Simulation experiment}\label{sec:application}

In this section we focus on the application of the CNN-IDE to SSTs. In Section~\ref{sec:data} we describe the data set used to train and test the IDE model; in Section~\ref{sec:estimation} we describe the estimation procedure; in Section~\ref{sec:forecasting} we provide a comparison study that assesses the performance of the CNN-IDE for probabilistic forecasting against other methods employing ST models. Finally, in Section~\ref{sec:radar} we show that the CNN-IDE trained on the SST data can be used (without parameter re-estimation) for nowcasting radar-reflectivity data, and remarkably perform comparably to other methods that involve estimating dynamical parameters using maximum likelihood.

Reproducible \texttt{R} code \citep{R} for the case studies shown in this section is available from \url{http://github.com/andrewzm/deepIDE}.

\subsection{Data}\label{sec:data}

\begin{figure}[t!]
	\includegraphics[width = \textwidth]{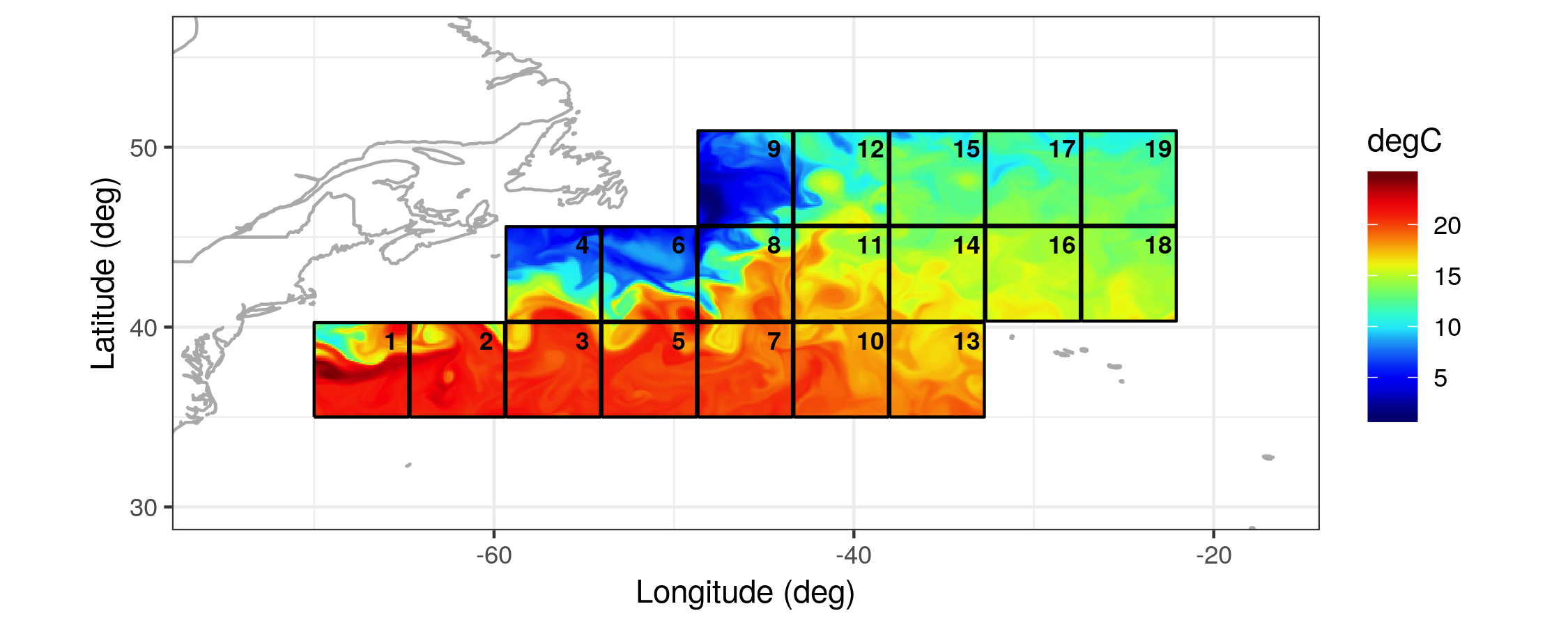}
	\caption{Sea-surface temperature in degrees Celsius from the CMEMS product \texttt{GLOBAL\_ANALYSIS\_FORECAST\_PHY\_001\_024} on 27 December 2006 in the North Atlantic Ocean. The boxes demarcate the 19 zones of size 64 $\times$ 64 used in our study.}\label{fig:zones}
\end{figure}

In this work we primarily focus on applying the CNN-IDE to modelling SST. We consider data available in the product \texttt{GLOBAL\_ANALYSIS\_FORECAST\_PHY\_001\_024} provided by the Copernicus Marine Environment Monitoring Service (CMEMS). This product contains daily means of several ocean-related variables such as temperature and salinity on a 1/12 degree lon--lat grid. As in \citet{deBezenac_2018} we consider daily SST from this product in the North Atlantic Ocean. Specifically, we consider 19 zones in this region, each of size 64 $\times$ 64 grid cells, and use the first 4003 days available in the product (i.e., from  27 December 2006 to 11 December 2017) for parameter estimation. The need to subdivide a large domain of interest into zones of manageable size is for computational reasons, but is not a significant drawback of the approach as long as the dynamics in the interior of a zone can be adequately captured by the most recent behaviour of the process in that zone. A disadvantage when predicting/forecasting is that a re-definition of zonal maps that considers overlapping zones and some post-hoc smoothing might be required to avoid boundary effects; we leave the consideration of such issues as future work. In our study we set $\tau = 3$ and have 19 zones, and therefore we have a total of 4000 $\times$ 19 = 76000 image sequences for maximum (conditional) likelihood estimation of $\psib$ and $\alphab$. Fig.~\ref{fig:zones} shows the 19 zone boundaries together with the SST product on 27 December 2006 within these zones. For convenience, when estimating the parameters, we map the spatial grid in each zone onto a $64 \times 64$ grid on the unit square. The chosen $\tau$ and the grid size were the largest we could have with the available memory on our GPU; these should be as large as possible. For dimension-reduction of the dynamics we  let $\{\phi_j(\cdot)\}$ in \eqref{eq:theta} be a set of $r = 64$ Gaussian radial basis functions regularly spaced in the unit square; in practice some model selection for an appropriate $r$ might be needed.

We modelled each zone independently but assumed that the CNN parameters $\psib$ are common across all the zones. We let the discrepancy term $\etab_t$ be zone-dependent, and therefore let the parameter vector $\alphab = (\alphab_1',\dots,\alphab_{19}')'$, where $\alphab_i \equiv (\sigma^2_i, \rho_i)'$ contains the variance and length scale of the covariance-function associated with the $i$th zone. In our application, we modelled the covariances using the Mat{\'e}rn covariance function with smoothness parameter $3/2$.  Considering the zones separately but allowing the residual variance and length scales to vary by zone is realistic when one considers that the fundamental dynamics should not change across the ocean, but that it is possible that the random forcing is zone-dependent (e.g., varying wind forcing across zones). 

Both the mean and variance of the SST within each zone vary by season and latitude. Seasonal effects can be included within the modelling framework and predicted and forecasted if desired.  However, this is beyond the scope of the analysis here and, for simplicity, we instead standardise the image pixels in each zone and time point by subtracting the average pixel value and dividing by the empirical standard deviation associated with that image. Modelling and inference is done using the standardised data that now (marginally) have approximately zero mean and unit variance; predictions and forecasts of the process are then unnormalised and reported on the original scale using the empirical averages and standard deviations that are assumed to be known.

The data that we use in this study come from a re-analysis and thus are complete and (can be assumed to be) noiseless. We therefore treat these data as \emph{process} data  (rather than \emph{observational} data), and use them for estimating $\psib$ and $\alphab$ directly through \eqref{eq:Lik}; we discuss parameter estimation in Section~\ref{sec:estimation}. We then use the CNN for prediction and forecasting on synthetically generated incomplete and noisy data in Section~\ref{sec:forecasting}, assuming that $\psib$ and $\alphab$ are fixed and known from their estimates. Reliable estimation of $\psib$ and $\alphab$ directly from observational data might be possible if these have high signal-to-noise ratio and are relatively complete, but such data is not always available. We provide further discussion on this point in Section~\ref{sec:discussion}.

\subsection{Parameter estimation}\label{sec:estimation}
 
We performed maximum likelihood estimation using the Adam optimiser together with the automatic differentiation facility in TensorFlow in \texttt{R} \citep{TensorflowR} on an nVIDIA 1080Ti GPU. We used a minibatch size of 16 on 90\% of the available data in a two-stage approach: We first estimated $\psib$ assuming that the elements in each $\etab_t$ are mutually uncorrelated, and subsequently estimated $\alphab$ from the fitted residuals. Note that we do not account for uncertainty in the CNN parameters; we suggest an approach that could remedy this in Section~\ref{sec:discussion}. When estimating $\psib$, convergence was assumed reached when the log-likelihood computed on the 10\% data left out for validation did not change substantially across two consecutive epochs; estimation of $\alphab$ is straightforward. We did not need any regularisation (such as dropout) to get a good fit, although we note that stochastic gradient descent tends to avoid spurious local maxima in the likelihood surface that often yield models that over-fit.

In our case, estimating the approximately 2 million parameters in $\psib$ and the 38 covariance function parameters in $\alphab$ required a few hours, with each minibatch log-likelihood and gradient evaluation requiring on the order of a tenth of a second. In total we used 30 epochs (i.e., each data sequence was used 30 times in total for log-likelihood evaluation).  Our maximum (conditional) likelihood estimates for $\{\sigma^2_i\}$ ranged between 0.003 and 0.024, while those for $\{\rho_i\}$ ranged between 0.034 and 0.049 (recalling that each zone was rescaled to the unit square). These estimates are indicative of non-trivial spatial residual variation.

\begin{figure}[t!]
  \includegraphics[width = 0.5\textwidth]{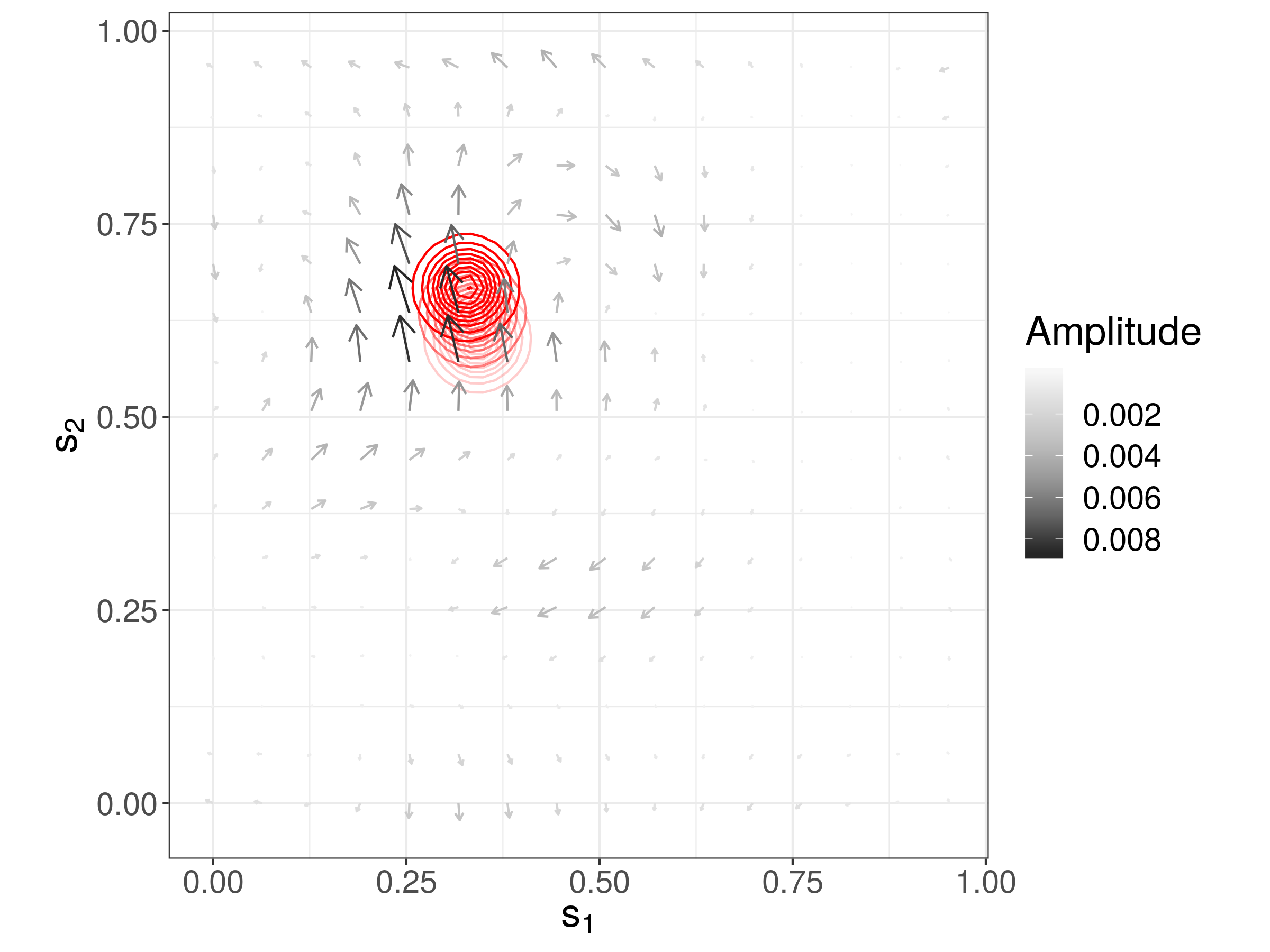}
  \includegraphics[width = 0.5\textwidth]{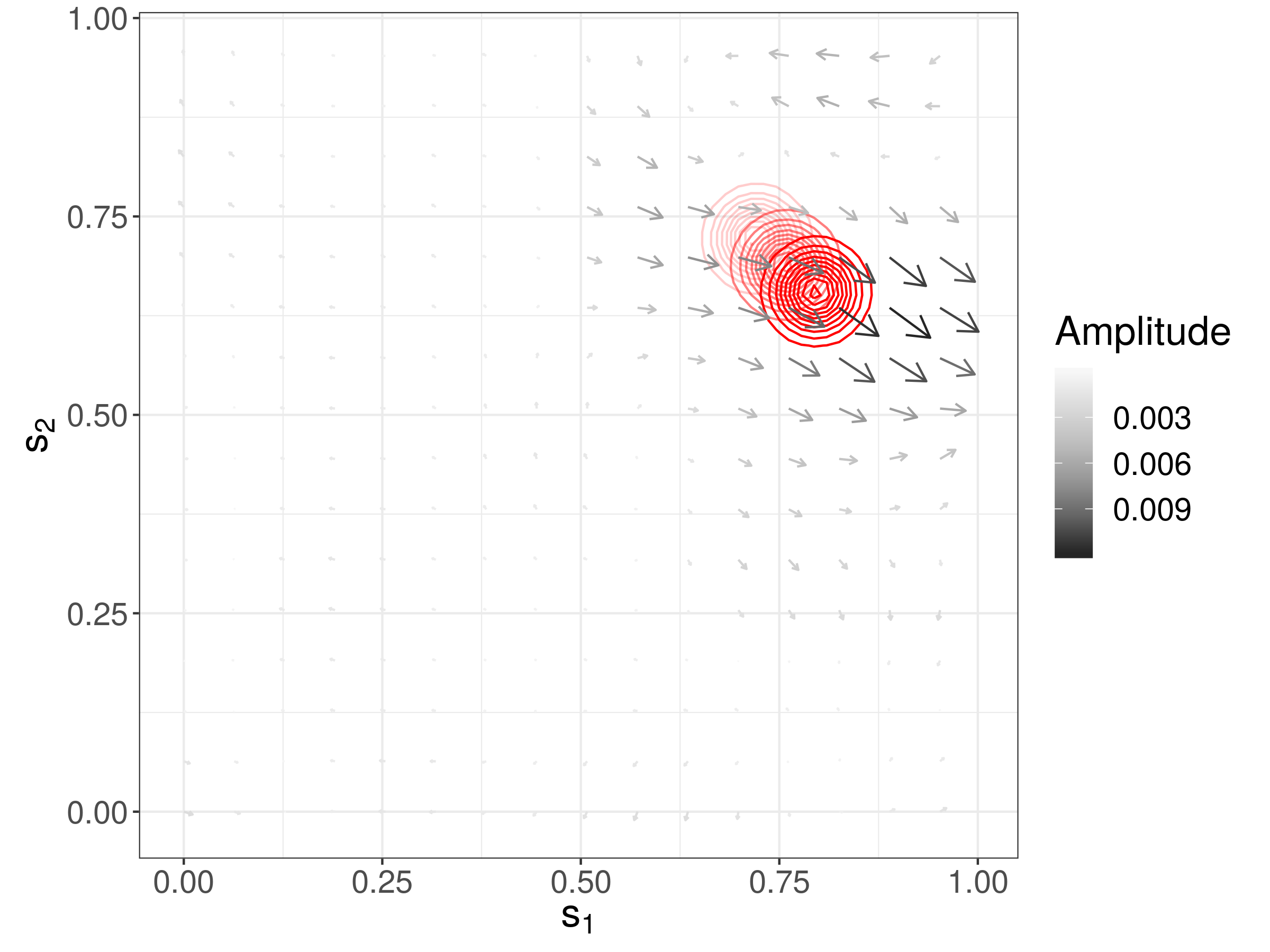}
  \caption{Output flow direction and intensity from the fitted CNN with $\tau = 3$ (arrows) in response to a Gaussian radial basis function shifting in time (contours), with decreasing transparency used to denote the function at times $t-2$, $t-1$ and $t$, respectively. The left and right panels show the outputs to two different input sequences.\label{fig:balls}}
\end{figure}

Interpreting the vector of estimated parameters $\hat\psib$ is not as straightforward. We can, however, get some insight into the behaviour of the fitted model by visualising the output of the CNN when a known input is supplied. In Fig.~\ref{fig:balls} we show the output of the flow (advection) parameters $\vartheta_{2}(\cdot)$ and $\vartheta_{3}(\cdot)$ on a domain $[0,1]\times[0,1]$ when the input is a Gaussian radial basis function shifting (spatially) in time. Note how the recovered flows are mostly localised, and broadly capture the direction of motion. The recovery is somewhat remarkable given that no (SST) training data was even vaguely similar to our test input in this example, and demonstrates that the fitted CNN is not over-fitted to the data, and is capturing the dynamics of the ST process as intended. We revisit the generality encapsulated by this global prior model in Section~\ref{sec:radar}.


\subsection{Comparative study}\label{sec:forecasting}

In Section~\ref{sec:estimation} we fitted the CNN and covariance function using directly observed, complete data from the SST product. In this section we show the model's use for forecasting in the realistic setting when observational data are incomplete and noisy. Here, the IDE plays the role of a statistical physical model in a data assimilation setting \citep[e.g.,][]{Wikle_2007} with the added advantage that it does not require any physical parameters to be estimated (e.g., diffusion parameters) online.

\begin{figure}[t!]
  \begin{center}
  \includegraphics[width = 0.75\textwidth]{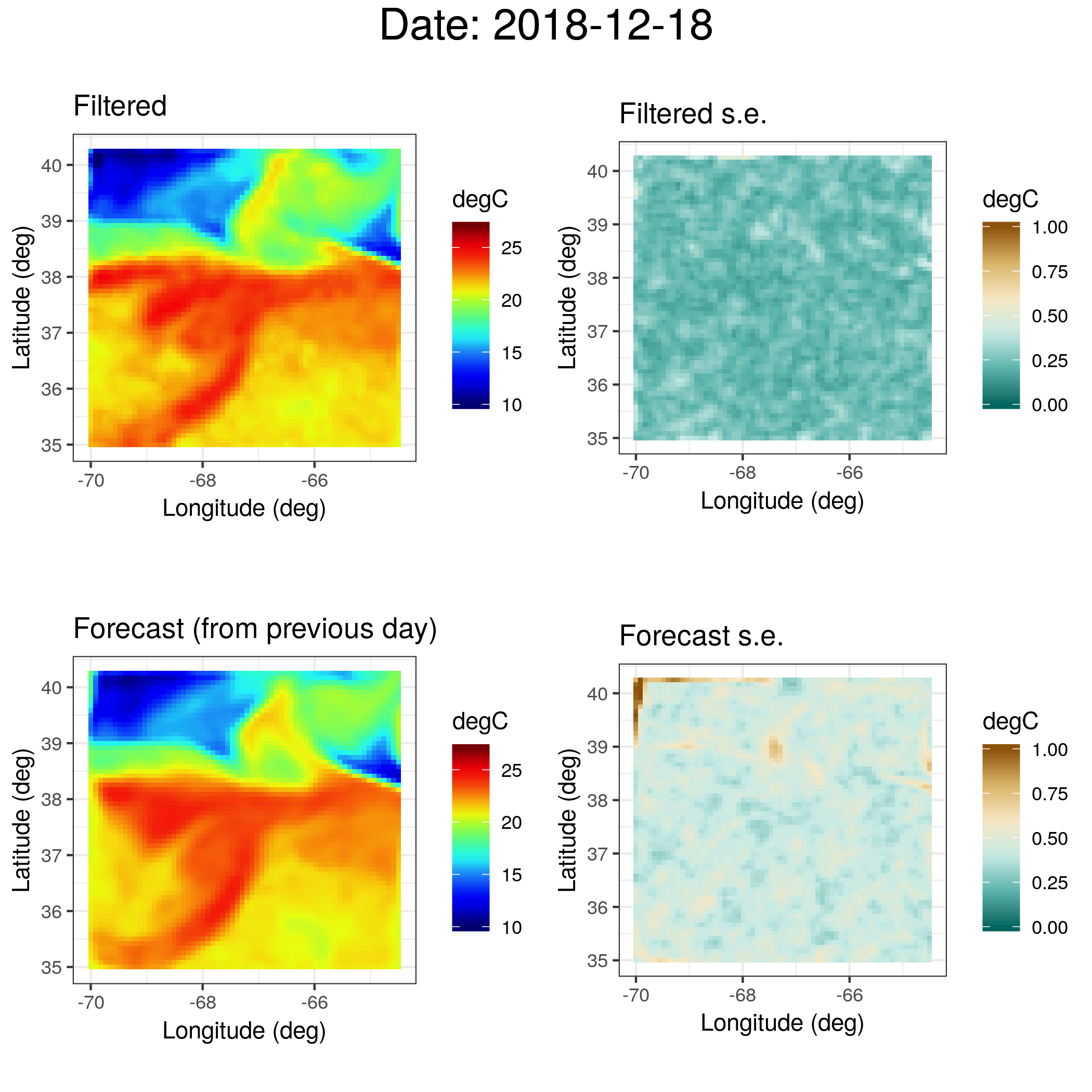}
  \caption{Screenshot from the animation showing the filtered and forecast estimates together with the associated uncertainty for the day of 18 December 2018. The animation is available from \url{https://github.com/andrewzm/deepIDE}.}\label{fig:screenshot}
  \end{center}
\end{figure}

We take the fitted CNN-IDE and use it for prediction and forecasting on data that has not been used for maximum-likelihood estimation or validation. Specifically, we take the standardised data in the same 19 zones from 01 September 2018 to 31 December 2018 and, for each day, we sample 1024 pixels at random, add Gaussian measurement error with known variance 0.01, and compute the filtered distributions of the process and its dynamical parameters, as well as the respective forecast distributions. In order to account for boundary issues, we only make inference, and compute diagnostics on, an interior domain of size 52 $\times$ 52 grid cells. An animation showing the filtered SST and dynamical field, along with associated parameters for one zone (Zone 1), is available from \url{https://github.com/andrewzm/deepIDE}. Fig.~\ref{fig:screenshot} shows a screenshot from this animation for the day of 18 December 2018. The spatial patterns of the filtered and forecast standard errors are largely driven by the measurement locations, which in this experiment change every day.

\begin{figure}[t!]
  \begin{center}
    \includegraphics[width = 0.49\textwidth]{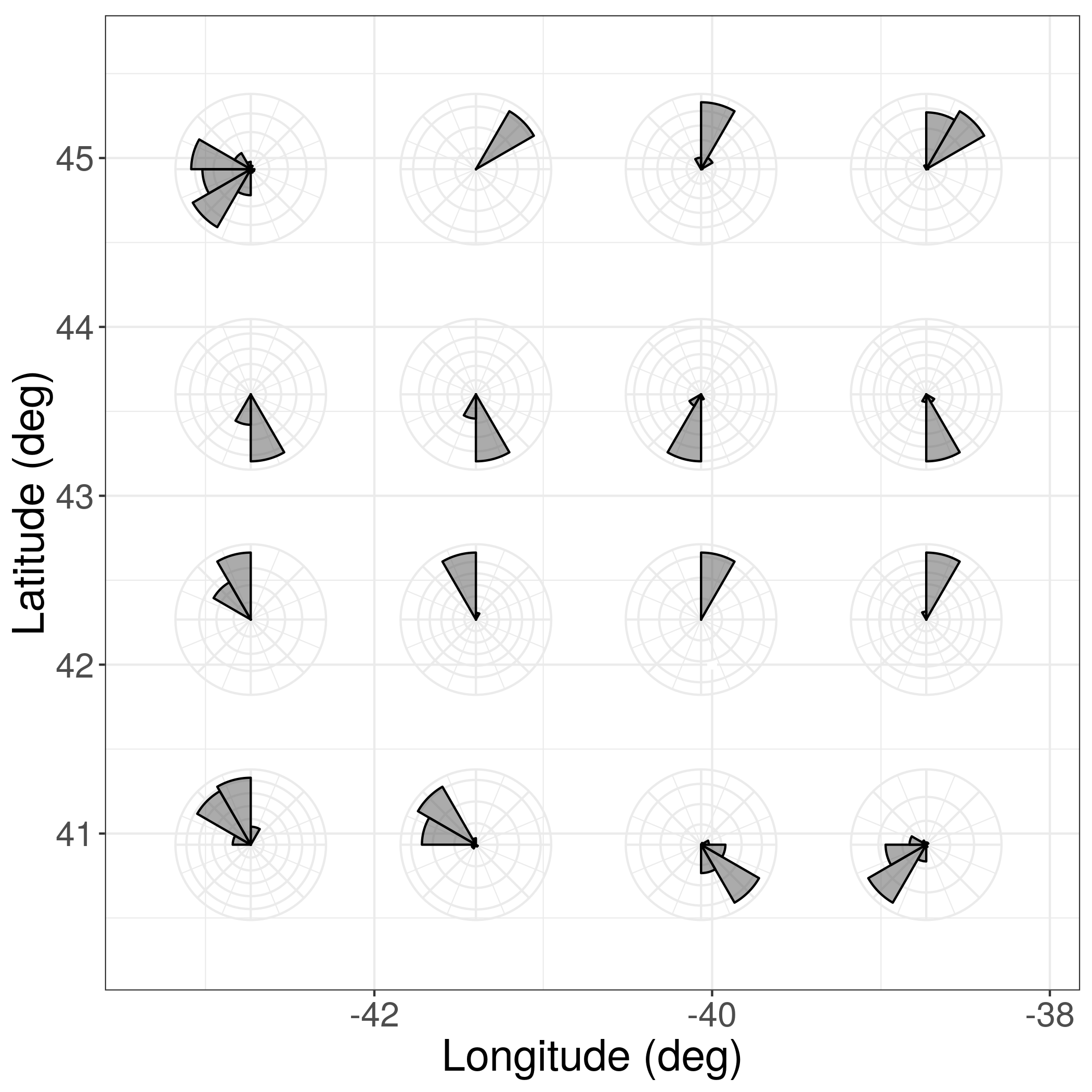}
    \includegraphics[width = 0.49\textwidth]{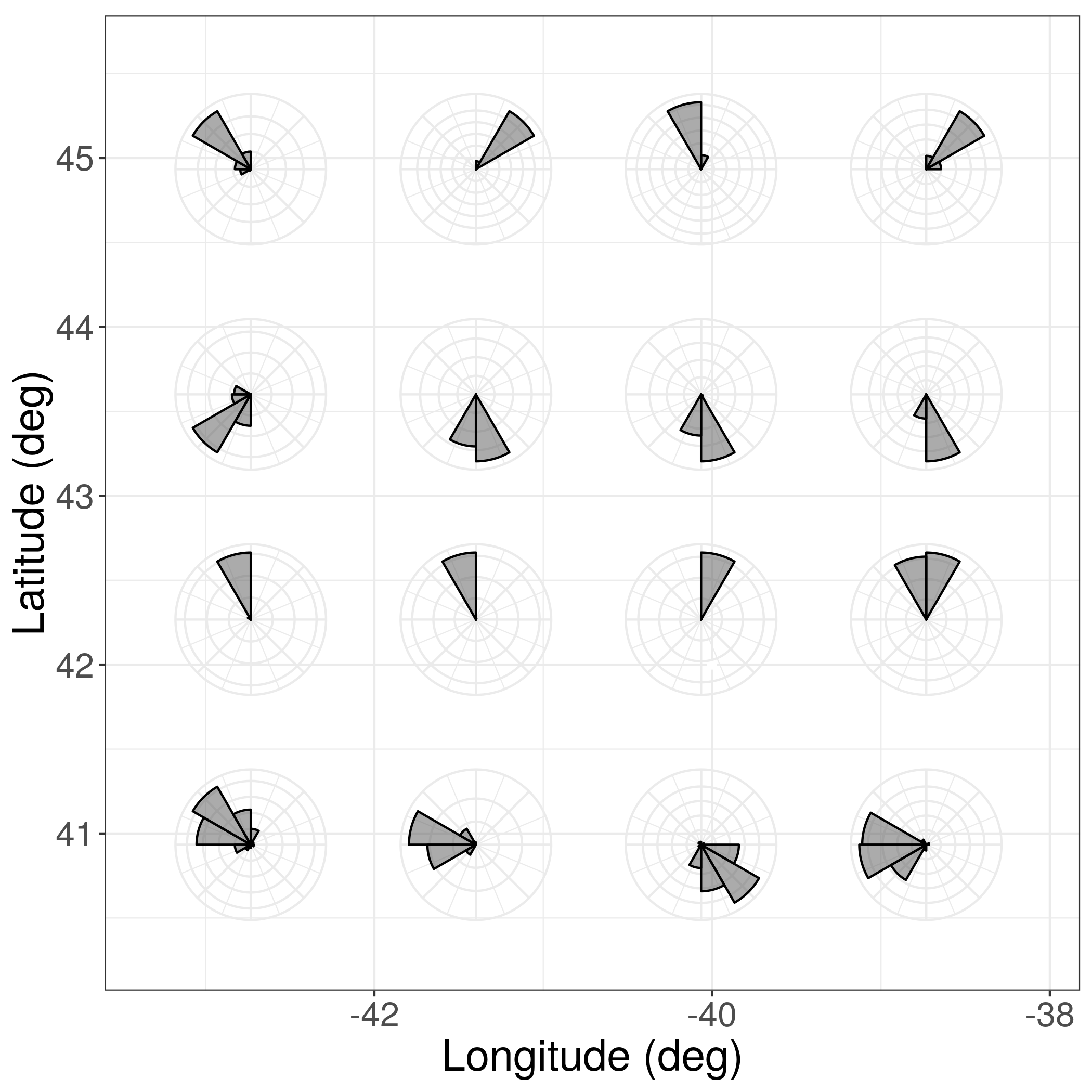}
  \caption{Unnormalised empirical histograms of the direction of process flow on 08 (left) and 09 (right) January 2007 at the shown locations on a 4 $\times$ 4 grid in Zone 11. The histograms were obtained by binning the CNN outputs from the 64 ensemble members, after the EnKF update step on the respective days, in bins of width $30^\circ$.}\label{fig:dynunc}
  \end{center}
\end{figure}

As discussed in Section~\ref{sec:inference}, we are also able to obtain filtered and forecast uncertainties on the process dynamics via the ensemble; in Fig.~\ref{fig:dynunc} we show unnormalised empirical histograms of the filtered flow directions within Zone 11 for 08 and 09 January 2007. Note how the predicted uncertainty is spatially variable, and how the dynamics do not fluctuate rapidly in time; this is expected from a slowly varying process (on a daily timescale) such as SST.

We compare these predictions and forecasts to those obtained from 

\begin{enumerate}
	\item Gaussian process regression (or simple kriging) of the data using a \emph{spatial} model consisting of an intercept, longitude and latitude as fixed effects (where the fixed effects are estimated using least squares and plugged in), and a Gaussian process with exponential covariance function. The measurement-error variance is fixed to the true variance, and for each time point a range parameter and a variance parameter are estimated using variogram techniques. Predictions at time $t$ are done using data at time $t$, while a na{\"i}ve forecast for time $t+1$ is done by simply reporting back the prediction at time $t$. Since SST evolves slowly over time, this forecaster does not perform too poorly, however it is to be expected that every model that takes time into consideration outperforms it. This forecaster thus plays the role of a baseline in our study. Model fitting and prediction were done using the \texttt{R} package \texttt{gstat} \citep{Pebesma_2004}.
	
	\item Gaussian process regression (or simple kriging) of the data using a \emph{spatio-temporal} model consisting of an intercept, longitude and latitude as fixed effects (estimated offline and plugged in), and a Gaussian process with exponential covariance function. The measurement-error variance is fixed to the true variance, and data between time $t - 3$ and time $t$ are used to make predictions at time $t$ and forecasts at time $t+1$. For each set of data points (across four time points), two range parameters (one for space and one for time) as well as a variance parameter, are estimated using maximum likelihood. Model fitting and prediction was done using custom GPU code via the R package \texttt{tensorflow}.
	
    \item The IDE model of \eqref{eq:IDE} combined with the data model \eqref{eq:datamodel} evaluated on the $64\times 64$ grid. This model is structurally identical to the CNN-IDE 
but now the kernel parameters are directly estimated from the data in a sliding window using maximum likelihood, and are not modelled to be state-dependent via the CNN. We fix the measurement-error variance to the true variance and use the data between time $t - 2$ and time $t$ to make predictions at time $t$ and forecasts at time $t+1$. Maximum-likelihood estimation was done using the innovations-form of the likelihood function \citep{Shumway_2006} via custom GPU code.  
 	
\end{enumerate}

\begin{figure}[t!]
  \begin{center}
  \includegraphics[width = \textwidth]{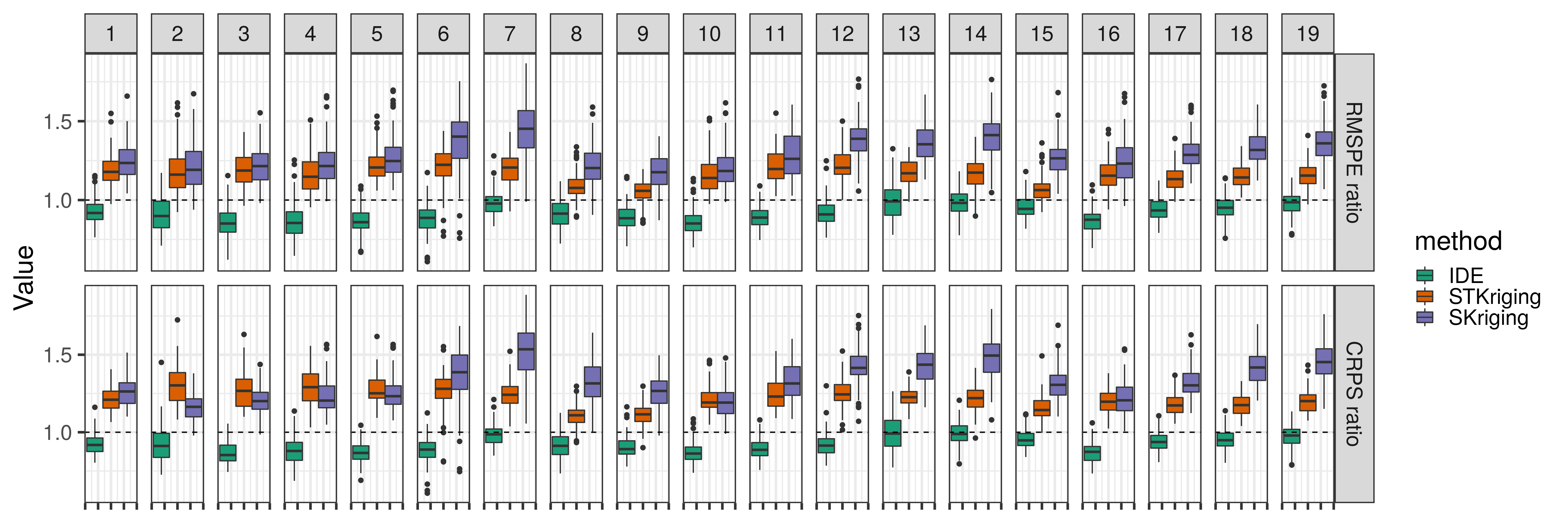}
  \caption{Box plots of the RMSPE ratio (top) and CRPS ratio (bottom) from daily filtered predictions made between September 2018 and December 2018 for the IDE with parameters estimated in a sliding window (IDE), spatio-temporal kriging (STKriging), and spatial kriging (SKriging) by zone (left to right). For each zone and time point, the ratio is computed by taking the diagnostic value associated with the respective method (IDE, STKriging, or SKriging), and dividing it by the value corresponding to the CNN-IDE. In each facet, the horizontal dashed lines marks the unit ratio (denoting identical performance). The boxes denote the interquartile range, the whiskers extend to the last values that are within 1.5 times the interquartile range from the quartiles, and the dots show the values that lie beyond the end of the whiskers. Values greater than 2 have been omitted for clarity.}\label{fig:filterresults}
  \end{center}
\end{figure}

\begin{figure}[t!]
  \begin{center}
  \includegraphics[width = \textwidth]{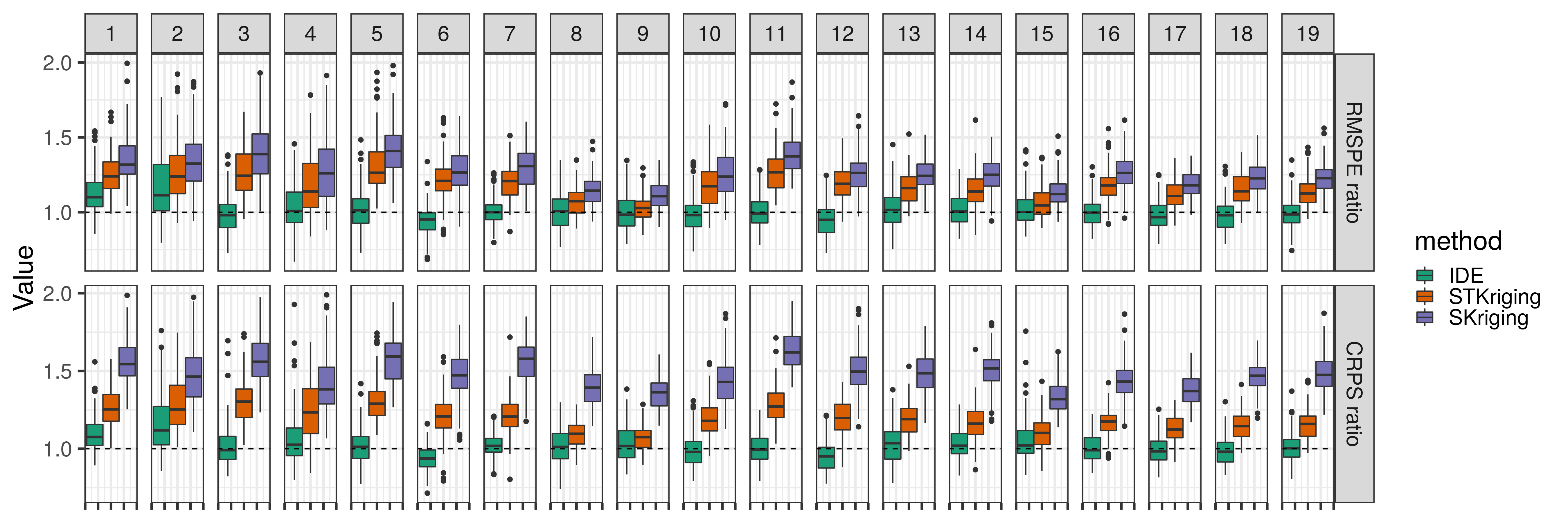}
  \caption{Same as Fig.~\ref{fig:filterresults} but for one-day-ahead forecasts. Recall that a forecast from spatial kriging at time $t + 1$ is simply the prediction at time $t$.\label{fig:forecastresults}}
  \end{center}
\end{figure}

In Fig.~\ref{fig:filterresults} we compare the root-mean-squared prediction error (RMSPE) and  continuous ranked probability score \citep[CRPS; see][]{Gneiting_2007} derived from the filtered distributions of the two kriging methods, the IDE model fitted using maximum likelihood at each time-point $t$, and the CNN-IDE model, in the 19 zones. Fig.~\ref{fig:forecastresults} is a comparison for the forecast distributions. As expected, we see that spatial kriging does not give filtered and forecast distributions that are unreasonable, but the methods that use ST models fare better, and the RMSPE and CRPS from the IDE models are always lower than those obtained with ST kriging. On the other hand, filtered predictions from the CNN-IDE are slightly worse ($\approx$ 10\%) in terms of RMSPE and CRPS than the same IDE model where the dynamical parameters are optimally re-estimated at each time step via maximum likelihood. This is an indication that the evolution of SST is approximately linear and time-invariant over the few days considered for the sliding window. Combined with the fact that the maximum likelihood estimator is asymptotically consistent \citep[][Chapter 7]{Caines_2018} and the Kalman filter is optimal for linear time-invariant Gaussian state-space models, suggests that the sliding window vanilla IDE is a gold standard when it comes to prediction in this application. We provide more discussion on why the CNN-IDE can be expected to perform slightly worse than the vanilla IDE, from a prediction perspective, in Section~\ref{sec:discussion}.

The main benefits of the proposed CNN-IDE are two-fold. The first benefit relates to the computational effort required to obtain the predictions and forecasts. 
The CNN-IDE was fitted offline in a few hours of computing time. Once fitted, it required only 20 minutes to do all the $\approx2000$ predictions and forecasts with 64 ensemble members. Spatial kriging required about 30 minutes while ST kriging about 12 hours.  The vanilla IDE required about 1 day of computing time since several gradient computations are required for estimating the parameters at each time point. This amount of time for computing was needed despite using optimised GPU code and the provision of reasonable initial values based on results supplied by the CNN. The CNN-IDE therefore provides a scalable way in which one can obtain reasonable predictions and forecasts on a global scale, where use of the vanilla IDE would be infeasible. The second benefit relates to model interpretation; the suite of fitted vanilla IDE models (one per time point and zone) only have local-space time interpretation. The CNN-IDE model, on the other hand, is a global model valid everywhere in space and time, and one that may also be used to forecast with other environmental processes that exhibit similar dynamics, as we show next.

\subsection{Applying the SST-trained IDE to radar-reflectivity data}\label{sec:radar}

In this section we carry out an unusual experiment, where we take the CNN-IDE with parameters estimated with the SST data and use it for forecasting radar reflectivity data. The data we consider are a set of 12 images of radar reflectivities obtained near Sydney, Australia, on 03 November, 2000, each corresponding to a 10-minute period, and regridded on a grid of size $64 \times 64$ grid cells. These data are supplied with the package \texttt{STRbook}, available from \url{https://github.com/andrewzm/STRbook}, and are described in more detail in \citet{Xu_2005}.

We estimated the measurement-error variance by considering an area of the first image with low precipitation, and taking the empirical variance of the pixel values there. This estimate was then plugged in the CNN-IDE and the vanilla IDE where parameters were estimated at each time step in a sliding window. As for $\alphab$ we used maximum likelihood estimates obtained from the SST experiment under the assumption that the discrepancy term is not zone-dependent. The EnKF was run using the CNN-IDE on the first 11 time points (i.e., from $t = 1$ to $t = 11$), and used to forecast the image at the final time point. Similarly, for the vanilla IDE, we used parameters estimated on data between $t = 9$ and $t = 11$ to forecast the image at $t = 12$.  We also compared these forecasts to those obtained by the package \texttt{IDE} which assumes linearity and temporal invariance over the entire time horizon, and which uses low-rank approximations to represent the field $\{Y_t(\cdot)\}$. Parameter estimation here was done using the first 11 radar images, and forecasts were provided for the image at the final time point.

In Table~\ref{tab:radar} we provide the RMSPE, CRPS, 90\% interval score, and the 90\% coverage for the forecasts from the three methods. We see that the CNN-IDE provides a forecast that is comparable to what can be obtained using the vanilla IDE with parameter estimation at every time step, and superior to what can be obtained using a low-rank version of the IDE. It also achieves nominal empirical coverage and gives low interval scores; this is despite it requiring a small fraction (on the order of a hundred times less) computation time required by the other methods. The low-rank IDE was very under-confident in its forecast, likely due to the presence of a large estimated fine-scale component of variation as a direct consequence of the low-rank approximation.
\begin{table}[t!]
  \caption{Forecasting diagnostics for the 12th radar-reflectivity image using data up to the 11th image for the different methods. Diagnostics shown are the root-mean-squared prediction error (RMSPE) in dB relative to Z (dBZ), the continuous-ranked probability score (CRPS), the 90\% interval score (IS90) and the 90\% coverage (Cov90). \label{tab:radar} \vspace{0.1in}}
  \centering
  
\begin{tabular}{lrrrr}
  \hline
Model & RMSPE (dBZ) & CRPS & IS90 & Cov90 \\ 
  \hline
CNN-IDE & 4.83 & 2.53 & 22.71 & 0.89 \\ 
  Full-rank IDE (window) & 4.89 & 2.54 & 24.09 & 0.89 \\ 
  Low-rank IDE & 5.72 & 4.99 & 60.53 & 1.00 \\ 
  \hline
\end{tabular}
\end{table}

\begin{figure}[t!]
  \includegraphics[width = \textwidth]{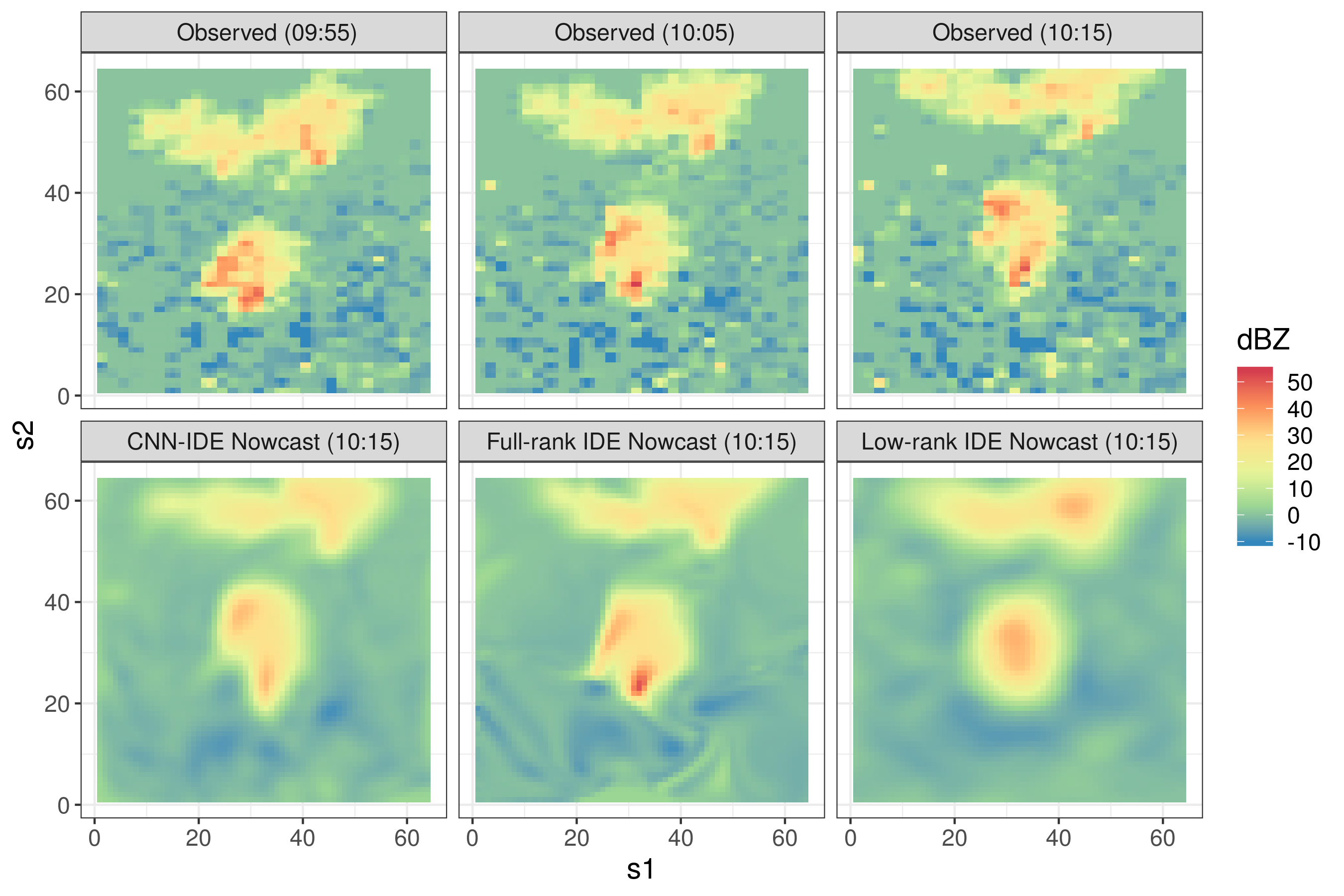}
  \caption{Top: Radar-reflectivity images at 09:55, 10:05, and 10:15 UTC (left to right, respectively). Bottom: 10-minute ahead nowcasts of the radar reflectivities obtained using the CNN-IDE, the full-rank vanilla IDE with parameter estimation in a sliding window, and a low-rank vanilla IDE through the \texttt{R} package \texttt{IDE} (left to right, respectively). \label{fig:radar}}
\end{figure}

In Fig.~\ref{fig:radar} we show the last three radar-reflectivity images in the sequence, and the 10-minute nowcasts of the final image from the three different methods. There is no clear difference between the nowcast of the CNN-IDE and the vanilla IDE where the dynamic parameters were estimated using data between $t = 9$ and $t = 11$. The low-rank IDE clearly captures the overall motion of the clouds apparent in the reflectivity images, but the forecast is over-smoothed as a result of the low-rank approximation.

\section{Discussion}\label{sec:discussion}

This work places the deterministic CNN model of \citet{deBezenac_2018} within a statistical hierarchical modelling framework. The resulting model allows us to consider noisy, incomplete measurements, and to provide filtered predictions as well as forecasts. Recasting the model as a statistical hierarchical model with a state-dependent kernel, we show how the ensemble Kalman filter can be used to concurrently quantify uncertainty in both the process and the dynamics. In our experimental study we found that the CNN-IDE was able to provide predictions and forecasts that are competitive with the vanilla IDE where parameters are estimated at each time step, and ones that are still superior to conventional ST kriging. The prediction and forecast uncertainties were also seen to be well calibrated.

Our results clearly show that if the system is approximately linear time-invariant over small time windows (as is the case with daily SST) there is no advantage, from a prediction performance point of view, in using a CNN-IDE over a vanilla IDE estimated in a sliding window. The slightly poorer predictive performance of the CNN-IDE could be due to a number of reasons. First, time-varying covariances in the random forcing, $\etab_t$, are not considered (while these are re-estimated in the sliding window). Second, while the CNN clearly extracts a mapping that is reasonable, this mapping is not perfect and results in some model misspecification. Third, the EnKF only yields an approximate filtering distribution when the dynamic equation is nonlinear (in this case state-dependent). 
However, we showed that the CNN-IDE encapsulates the dynamical behaviour of the system, so that prediction and forecasting can be done with very little or no parameter estimation, and very quickly using inexpensive computing hardware. We showed that the trained CNN-IDE can even be used for modelling and forecasting entirely different environmental processes. This approach for ST forecasting is hence very versatile and scalable.

The simulation experiment of Section~\ref{sec:application} considered noisy data generated from the same analysis product used to train the model (albeit at different time points). Here, the component of variation $\eta_t(\cdot)$ captures the discrepancy between the CNN-IDE and the analysis product it is fit to.  When observational data are used, a second component of variation may be needed to account for additional discrepancies that exist between the analysis product and the true process \citep{Brynjarsdottir2014learning}. This component of variation, as well as measurement error, can be determined offline, as we did in Section \ref{sec:radar}. Alternatively, the ensemble Kalman filter may be placed within a parameter estimation framework \citep[e.g.,][]{Katzfuss_2019}, wherein the variance components are estimated. Even in a framework where parameter estimation is needed, use of the CNN-IDE may still be beneficial since it precludes estimation of the dynamical parameters, which are sometimes difficult to estimate in both Bayesian and maximum-likelihood settings.

We have assumed that $\psib$ is fixed and unknown. However, despite its high dimensionality, one can place prior distributions on $\psib$. The posterior distribution over $\psib$ is analytically intractable, but approximate inference methods such as stochastic variational Bayes \citep{Zammit_2019}, can readily be adapted to this scenario. One may also entertain the idea of updating $\psib$ \emph{online}, that is, using the observations. This is worth investigating if, for example, abundant high quality satellite imagery is available. In this case, $\hat\psib$ initialised using the analysis product would serve as a suitable initial value which is then updated. Estimating $\psib$ from a product or inventory, and using those estimates within a data assimilation framework is not uncommon; see, for example, \citet{leeds2014emulator}, \citet{Zammit_2015b} or \citet{Zammit_2016} and references therein for similar modelling strategies.

In our simulation experiment in Section~\ref{sec:application} the CNN was trained on tens of thousands of spatial images, while all other methods only had a handful of spatial images on which to conduct parameter estimation, and thus had to be relatively parsimonious. This comparison, however, serves to highlight an important caveat of statistical ST models in common use today, namely, that there are a limited number of parametric options available (e.g., quadratic nonlinear models) that are sufficiently flexible to harness the complexity of the dynamics that can be present in many environmental process.  Deep neural nets, and CNNs in particular, contain the required structure to harness this complexity. The requirement that they need a relatively huge amount of data to fit is rather benign in today's world where several scientific domains benefit from considerable amounts of satellite and model output data.

Future work will endeavour to propagate uncertainties of the CNN parameters to those on the predictions, and to uncover other application areas, potentially with non-Gaussian data, nonlinearity, and with a change-of-support problem, which can benefit from this type of modelling framework.  In addition, the consideration of alternative computational approaches to allow the method to be applied to higher-dimensional spatial fields is an area of future research.

Finally, the work in this article highlights the important role models that are commonly employed in machine learning can play in geostatistics, and statistics at large \citep{wikle2019comparison}. As we show here, the common criticism that they are not designed to handle uncertainty can be mitigated to a large extent with the use of hierarchical statistical frameworks. Another criticism that they are overly complex and difficult to interpret is offset by our ability to investigate the mechanistic information they encode (e.g., Fig.~\ref{fig:balls}) and their potential for competitive predictive/forecasting performance at a fraction of the computational cost. It is likely that the next few years will see machine learning models, and in particular deep neural nets, revolutionising the field of spatio-temporal statistics as we know it.

\section{Acknowledgements}

AZ-M's research was supported by the Australian Research Council (ARC) Discovery Early Career Research Award, DE180100203.  CKW was supported by the National Science Foundation (NSF) Award DMS-1811745.

\bibliography{Bibliography}

\end{document}